\documentclass[a4paper,fleqn]{cas-dc}

\usepackage[numbers]{natbib}
\usepackage{multirow}
\usepackage{stfloats}
\usepackage{multirow}
\usepackage{color}
\usepackage{amsmath}
\usepackage{graphicx}
\usepackage{url}
\usepackage{amssymb}
\usepackage{booktabs}
\usepackage{subfigure}

\def\tsc#1{\csdef{#1}{\textsc{\lowercase{#1}}\xspace}}
\tsc{WGM}
\tsc{QE}
\tsc{EP}
\tsc{PMS}
\tsc{BEC}
\tsc{DE}

\begin{document}
\let\WriteBookmarks\relax
\shorttitle{Multiple Attentional Pyramid Networks for Chinese Herbal Recognition}
\shortauthors{Yingxue Xu}

\title [mode = title]{Multiple Attentional Pyramid Networks for Chinese Herbal Recognition}                      
\tnotemark[1]

\tnotetext[1]{This study was supported by China National Science Foundation (Grant Nos. 60973083 and 61273363), Science and Technology Planning Project of Guangdong Province (Grant Nos. 2014A010103009 and 2015A020217002), and Guangzhou Science and Technology Planning Project (Grant No. 201604020179, 201803010088).}

\author[1]{Yingxue Xu}[style=chinese]
\ead{csxuyingxue@mail.scut.edu.cn}
\address[1]{School of Computer Science \& Engineering, South China University of Technology}

\author[1]{Guihua Wen}[style=chinese]
\cormark[1]
\fnmark[1]
\ead{crghwen@scut.edu.cn}
\author[1]{Yang Hu}[style=chinese]
\fnmark[1]


\author[1]{Mingnan Luo}[style=chinese]

\author[1]{Dan Dai}[style=chinese]
\author[1]{Yishan Zhuang}[style=chinese]
\author[2]{Wendy Hall}
\ead{wh@ecs.soton.ac.uk}
\address[2]{ Web Science Institute, University of Southampton, UK}
\cortext[cor1]{Corresponding author}

\fntext[fn1]{Guihua Wen and Yang Hu contributed equally to this work.}

\begin{abstract}
Chinese herbs play a critical role in Traditional Chinese Medicine. Due to different recognition granularity, they can be recognized accurately only by professionals with much experience. It is expected that they can be recognized automatically using new techniques like machine learning. However, there is no Chinese herbal image dataset available. Simultaneously, there is no machine learning method which can deal with Chinese herbal image recognition well.
Therefore, this paper begins with building a new standard Chinese-Herbs dataset. Subsequently, a new Attentional Pyramid Networks (APN) for Chinese herbal recognition is proposed, where both novel competitive attention and spatial collaborative attention are proposed and then applied. APN can adaptively model Chinese herbal images with different feature scales. Finally, a new framework for Chinese herbal recognition is proposed as a new application of APN. Experiments are conducted on our constructed dataset and validate the effectiveness of our methods.
\end{abstract}

\begin{keywords}
Pyramid networks, Attention mechanism,  Multi-scale features, Chinese herbal recognition, Chinese herbs image datasets.
\end{keywords}

\maketitle

\section{Introduction}
As a simple and non-invasive treatment with the minor side effect, Traditional Chinese Medicine (TCM) plays a significant role in health-care for several thousand years. It is widely used in China and numerous Asian countries~\cite{Jinghua2017, zuo2018network}. Therefore, there are lots of researches on TCM investigated to perform auxiliary diagnosis and treatment using the advanced techniques, such as image processing methods~\cite{Zhang2014} and newly proposed deep learning methods~\cite{Wen2019a}. Chinese herbs, as a part of TCM, have excellent performance on not only the treatment of the diseases but also the physical conditioning with the guidance of the theoretical system of TCM, so that they have gradually become a part of people's life. However, due to a lack of professional equipment and knowledge, it is difficult for non-professionals to recognize Chinese herbs accurately so that Chinese herbs recognition tools are heavily expected. However, to the best of our knowledge, there is no research on this task currently. As there is no available Chinese herbal image dataset at present, Chinese herbal recognition seems much tricky, even if there are many excellent machine learning methods available, such as excellent deep convolution neural networks (CNNs) for visual object recognition \cite{qi2019exploiting} and detection \cite{zhang2018weakly,zhang2019hyperfusion,li2019illumination}. Therefore, this paper begins with constructing a new Chinese herbal image dataset.

\begin{figure*}
\label{fig_exhibition}
\centering
\includegraphics[scale=0.55]{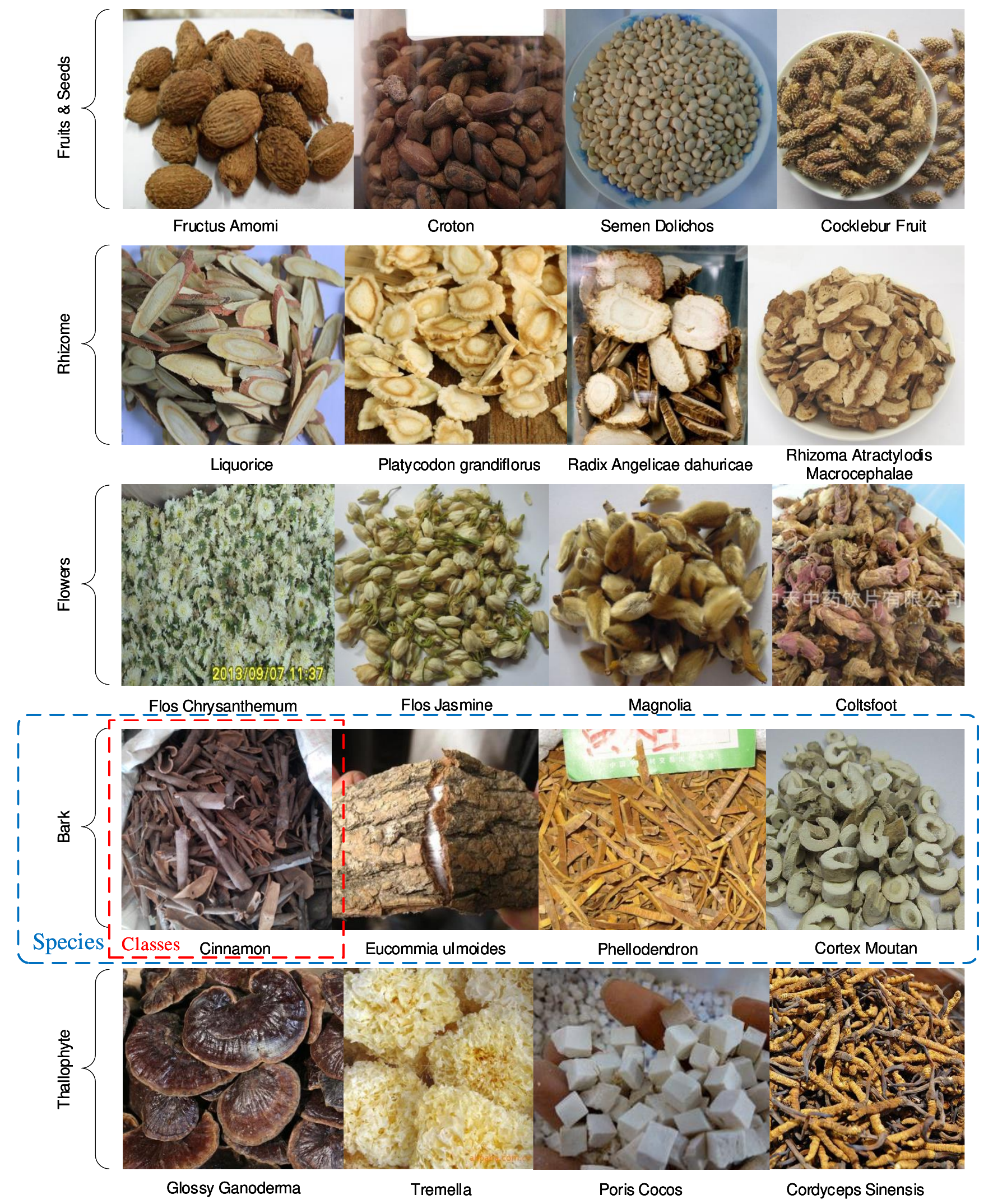} 
\caption{Examples of main species in CNH-98. The {\color[RGB]{24,116,205}{blue}} box represents an example of the \textit{Species} and the {\color{red}{red}} box indicates an examples of the \textit{Class}.}
\end{figure*}

\begin{figure}[b]
\centering
\includegraphics[scale=0.35]{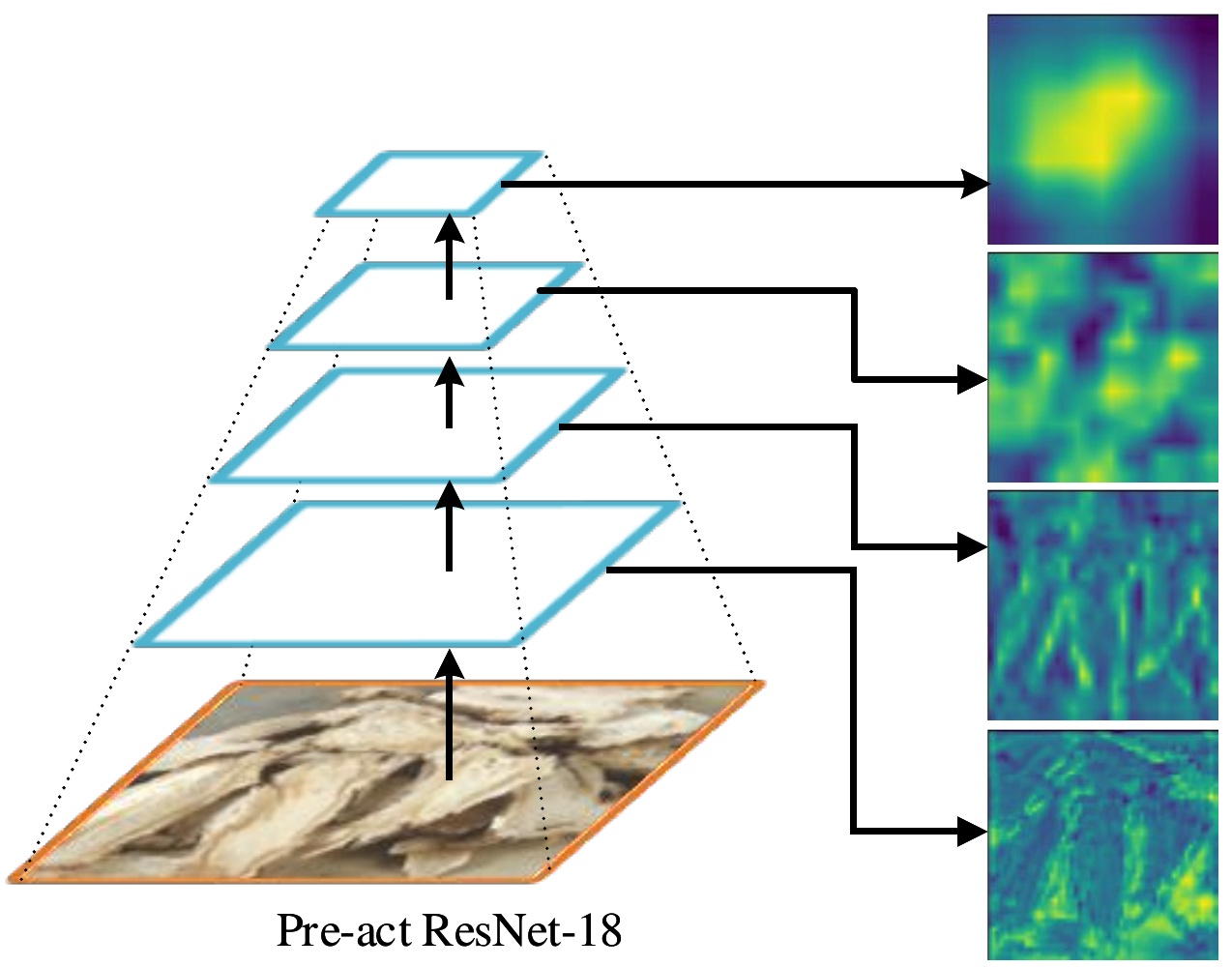}
\caption{Feature maps from various layers of CNN, where feature maps are extracted from four levels of ResNet-18 to form a feature pyramid. It can be seen that the features from different layers of networks are diverse.}
\label{fig_feature_pyramid}
\end{figure}
\indent
Chinese herbal recognition differs from both the regular image recognition ~\cite{zhu2019distance} and the fine-grained image recognition~\cite{qi2019exploiting}, where the former focuses on global semantic information such as the outline and shape, while the latter needs the more detailed local information. Chinese herbal recognition is confronted with two problems: (1) Some Chinese herbs, belonging to different species, are easy to be classified by the global shape features, as shown in Figure  \ref{fig_exhibition}. For example, Fructus Amomi and Liquorice are much different in global shape so that they can be easily classified using the shape features. (2) There are some other Chinese herbs with similar shapes belonging to the same species, requiring the more fine-grained features for the recognition, such as Croton and Cocklebur Fruit in Figure  \ref{fig_exhibition}, which have only a few differences in details. Therefore, the recognition granularity of inter-species differs from intra-species, where the former is easier to be classified than the latter. Thus we convert Chinese herbal recognition to a multi-granularity task. Generally, it is hard to take into account the above two cases when we only use the single-scale features, as the regular CNNs cannot detect the scale adaptively. Therefore, we need to consider multi-scale features and adaptively emphasize the corresponding scales that are efficient for different herbs, since features with different scales indicate the different granularity.\\
\indent
If CNNs are used, the features extracted from the convolution layers of different depth are rich in diversity. It is empirically confirmed that the features of the shallower layer contain both more details and spatial information due to the smaller convolution receptive field and high resolution, while the features of the deeper layer are more global and semantic in contrast~\cite{yu2018deep}. They can be illustrated in Figure \ref{fig_feature_pyramid} and the representation visualization~\cite{yu2017dilated}. To consider these two cases simultaneously, the pyramid structure of Feature Pyramid Networks \cite{lin2017feature} (FPN) can naturally leverage ConvNet's pyramid feature hierarchy to aggregate features with different scales from a variety of levels. Namely, the pyramid structure of FPN can fuse semantic and spatial information \cite{yu2018deep} required by multi-granularity recognition task like Chinese herbal recognition.\\
\indent
FPN is designed for object detection, which attaches predictors to all RoIs of all levels. However, when it is applied to the recognition task that requires multi-scale features, we have to feed the features of all levels into one classifier. In this way, it is much possible to generate a large number of redundant features, making it difficult to optimize the network during training. In such a case, some efficient features are expected to be selected. On the other hand, FPN implies a strong constraint that features from the adjacent levels are combined in a fixed way, which easily leads to inefficient fused features for samples with different granularity. Furthermore, its networks cannot dynamically adjust weights of features from different levels. In order to overcome these disadvantages, this paper proposes a new Attentional Pyramid Networks (APN) for Chinese herbal recognition, where both novel Competitive Attention (CA) and novel Spatial Collaborative Attention (SCA) are proposed and then applied. In this way,  APN can obtain the more efficient fused features and adaptively model Chinese herbal images with different feature scales.\\
\indent
As a kind of channel-wise attention mechanism based on SENet~\cite{hu2019squeeze}, CA can automatically select features of different levels for different samples in a soft way when aggregating features of different levels. SE in SENet is limited to re-scaling weights of features from the single layers and single information stream, while the proposed CA extends the modelling ranges of channel-wise attention and spatial attention to multiple streams. It explicitly models the channel dependencies in the process of across-level features fusion, indicating an implicit competition between spatial and semantic information streams. On the other hand, channel-wise attention solely considers relatively the global information for each feature map, but we need to dynamically trade off the global and local information across different levels. Local details are usually reflected by the spatial dimension, so that the proposed SCA is based on the spatial attention~\cite{Li2018Harmonious} by collaboratively learning from information streams of different levels. It can collaboratively trade-off the global and local information of features from different levels and mutually complement for each other in the process of spatial attentional modelling. The competition and trade-off can help the pyramid network produce fused features that are more efficient for the multi-granularity task.
SCA has two versions. One is parameter-free, which solely exploits the global descriptor of each channel to model spatial relations and ignores spatial clues between various channels. However, it is generally known that feature maps of the different channels tend to focus on the different spatial regions. Some pay attention to the global information such as the background and shape, while others to the local details like texture, which can be observed from the representation visualization \cite{yu2017dilated}. Therefore, there may be a certain spatial correlation between different channels~\cite{wu2018group}, which should not be neglected. Parametric SCA can be applied to deal with this issue, which can spatially capture the global and local complementary relationship between a variety of channels achieved by parametric filters. Obviously, both CA and SCA are easily extended to other neural networks with feature fusing on multiple information streams. Also, APN can be generalized to similar multi-granularity recognition tasks. In summary, several novel contributions are presented as follows:
\begin{enumerate}
  \item A new standard Chinese-Herbs dataset is constructed, which not only supports Chinese herbal recognition but also supports the research of machine learning methods.
  \item A new Attentional Pyramid Networks (APN) for Chinese herbal recognition is proposed, where both novel competitive attention and novel spatial collaborative attention are proposed and then applied. APN can obtain the
more efficiently fused features and adaptively model Chinese herbal images with different feature scales.
  \item The proposed CA can selectively place different emphasis on the features of different level for the multi-granularity task whereas SCA can adaptively quantify the importance of each image region. Furthermore, two variants of SCA are proposed to obtain the more effective performance.
   \item As an application, a new framework with APN for Chinese herbal recognition is proposed. Experimental results on our constructed dataset validate the effectiveness and the superior performance of our methods.
\end{enumerate}

\section{Related Work}
\textbf{Feature Pyramid.} Currently, a variety of methods~\cite{huang2018multi-scale} are proposed to process multi-scale spatial information by aggregating multi-scale features.~\cite{yu2018deep} presents a summary of the typical feature aggregation methods such as feature pyramid \cite{lin2017feature} applied on various fields to handle multi-scale tasks, such as segmentation \cite{kirillov2019panoptic,chen2018deeplab}, object detection \cite{lin2017feature}, and pose estimation \cite{chen2018cascaded}. However, there are fewer multi-scale methods for image recognition task~\cite{he2015spatial}. Definitely, they have not been applied to Chinese herbal recognition.

Generally speaking, there are two primary approaches to exploit multi-scale features: pyramid pooling~\cite{long2015fully} and encoder-decoder architecture~\cite{lin2017feature} achieved by skip connections. As for the former, ParseNet~\cite{Liu2015ParseNet} introduces global context informations to FCN~\cite{long2015fully}. Moreover, DeepLab v2~\cite{chen2018deeplab} embeds multi-scale features by spatial pyramid pooling for image segmentation, based on parallel dilated convolutions.

Pyramid encoder-decoder architecture is proposed to combine spatially strong features with semantically powerful features achieved by skip-connection. UNet~\cite{ronneberger2015u} can aggregate coarse-to-fine features for biomedical images. Furthermore, RefineNet~\cite{lin2017refinenet} follows this main idea. Besides, a novel encoder-decoder architecture, the hourglass architecture~\cite{newell2016stacked}, stacks multiple encoder-decoder structures on each block. The hourglass structure is further applied on residual blocks to propose a pyramid residual module.~\cite{yang2017learning} and~\cite{ning2018knowledge} further introduce prior knowledge into it. Inspired by hourglass blocks, FPN~\cite{lin2017feature} designs a novel hourglass pyramid network with strong semantic at all scales by top-down pathway and lateral skip connections and RON~\cite{kong2017ron:} also use a similar idea implemented by reverse connections. Based on FPN, Panoptic FPN~\cite{kirillov2019panoptic} propose a more complex feature hierarchical architecture for panoptic segmentation. However, these approaches did not use attention mechanism. Our method proposed and then applied two novel attention model: competitive attention and spatial collaborative attention.

\textbf{Attention in CNN.} The attention mechanism has been applied to the modelling process of CNNs~\cite{Nguyen2018Attentive}, which primarily involves two aspects: channel-wise attention mechanism~\cite{hu2019squeeze} and spatial attention mechanism~\cite{wang2017residual, Li2018Harmonious}. The former explicitly captures interdependency between channels and the other one re-weights the spatial signals of images. Furthermore, some models combine both spatial and channel-wise attention, such as SCA-CNN~\cite{Chen2017SCA} and CBAM~\cite{woo2018cbam}. However, the mentioned models are limited in the local region. To solve this problem, self-attention~\cite{wang2018non-local} is proposed to capture long-range dependencies between local and global information. Interaction-aware pyramid~\cite{du2018interaction-aware}, a self-attention model, introduces attention into the pyramid network for modelling long-range relationship.
Additionally, there are some attention models based on domain knowledge~\cite{Chen2017Attentive, Choi2017GRAM}. Different from these attention methods, based on the structure of pyramid networks, our proposed attention mechanism explicitly models on multiple information streams. Namely, it takes into consideration the interaction of different information streams and joint modelling, which considers both the implicit competition and trade-off simultaneously.

\textbf{CNN Applied on Tasks like Herbal Recognition.} There are some similar tasks finished by CNN, such as plants recognition~\cite{toth2016deep}, leaf recognition~\cite{hu2018multiscale}, flower recognition~\cite{Gogul2017Flower}. However, to the best of our knowledge, there is no research on Chinese herbal recognition using CNN, let alone APN.

\section{Proposed method}
\begin{figure*}
\includegraphics[scale=0.44]{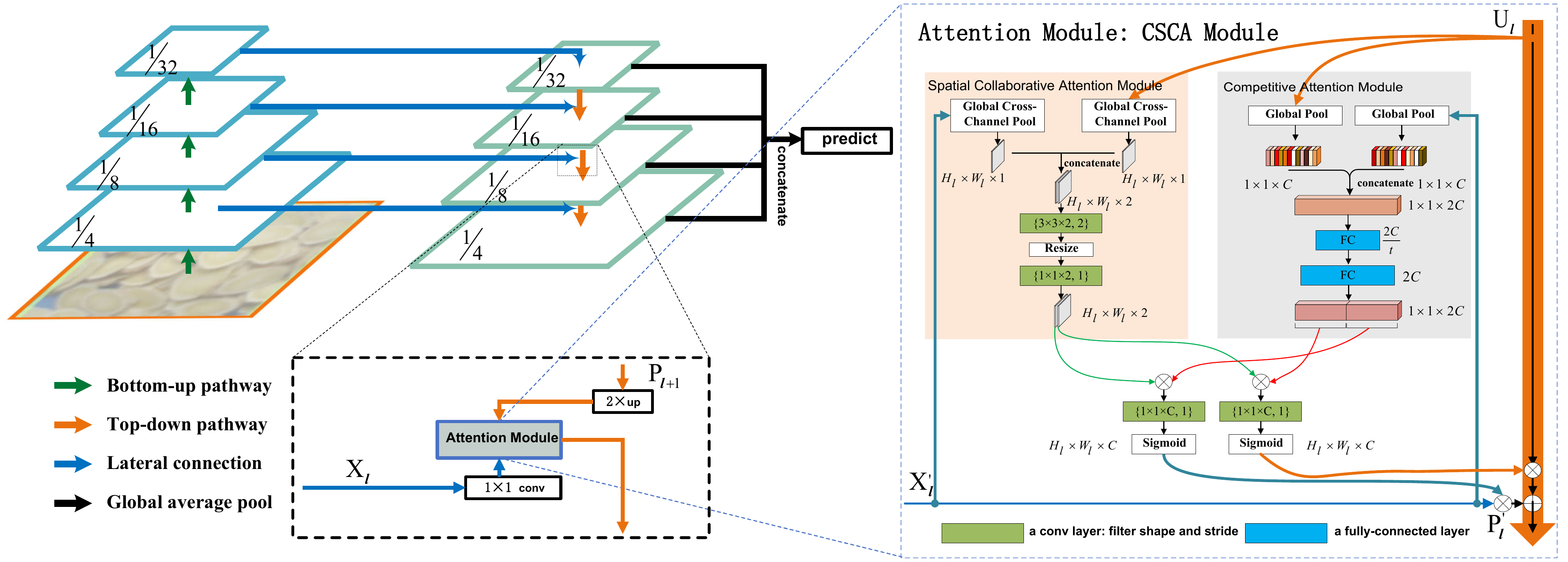}
\caption{Overview of Attentional Pyramid Network and its Attention modules: Competitive Attention and Spatial Collaborative Attention, where Batch Normalisation~\cite{ioffe2015batch} is used before sigmoid but not shown for brevity. All feature maps are resized in SCA module by the bilinear interpolation (default).}
\label{fig_4_1_2}
\end{figure*}

The overview of the proposed Attentional Pyramid Networks (APN) is shown in Figure  \ref{fig_4_1_2}, where the architecture of the pyramid network is presented on the left and its attention module on the right.
\subsection{Pyramid Networks in APN}
\label{sec:overview of fpn}
The pyramid networks in APN is innovated from FPN~\cite{lin2017feature}. Following FPN, it constructs a standard neural network as its backbone with multi-scale features, where the spatial resolution of each level is reduced by 50\% from the previous level on the bottom-up pathway. Here we record the output of $l$-th level on the bottom-up pathway as $\mathbf{X}_l \in \mathbb{R}^{H_l\times W_l \times C_l}$. The features $\mathbf{X}_l$ would be transformed into ${\mathbf{X}'}_l \in {\mathbb{R}^{H_l\times W_l \times C}}$ by lateral connections achieved by a convolution operation with the filter size of $1\times 1$.

Starting from the coarsest-resolution feature maps ${\mathbf{X}'}_L$ that locate at the highest pyramid level $L$, the top-down pathway progressively generates higher resolution features $\mathbf{U}_{l} \in \mathbb{R}^{H_{l}\times W_{l} \times C}$ by upsampling the output $\mathbf{P}_{l+1} \in \mathbb{R}^{H_{l+1}\times W_{l+1} \times C}$ of higher-level $l+1$ by a factor of 2. Namely, given an intermediate feature map with the size of $H_{l+1}\times W_{l+1}$ as the input of a level, top-down pathway sequentially infers a feature map with the size of $2H_{l+1}\times 2W_{l+1}$ as the output of this level. Then, these upsampled features $\mathbf{U}_{l}$ would be merged with the transformed features ${\mathbf{X}'}_l$ by element-wise addition:
\begin{equation}
\label{equ:fpn_pl'}
\mathbf{P'}_l ={\mathbf{X}'}_l+\mathbf{U}_{l},\quad\text{if}\ l=1,2,\ldots,L-1
\end{equation}
After smoothed by a convolution operation with the filter size of $3\times 3$ to reduce aliasing effect of upsampling, $\mathbf{P'}_l$ would be transformed into $\mathbf{P}_l \in \mathbb{R}^{H_{l}\times W_{l} \times C}$ as the output of level $l$. As a result, the output of each pyramid level on top-down pathway can be concluded as:
\begin{equation}
\label{equ:fpn_Pl}
\begin{cases}
\mathbf{P}_L={\mathbf{X}'}_L, & \text{if}\ l=L \\
\mathbf{P}_l=f^{3\times 3}(f^{1\times 1}(\mathbf{X}_l)+F_{up}(\mathbf{P}_{l+1})), & \text{if}\ l <L\\
\end{cases}
\end{equation}
where we denote $f^{k\times k}$ as the convolution operation with the filter size of $k\times k$ and $F_{up}$ represents the upsampling operation.

As opposed to the transformed feature $\mathbf{x}'_l$ from the bottom-up pathway, $\mathbf{u}_{l}$ are spatially coarser but semantically stronger. Hence we naturally refer the bottom-up pathway to spatial flow and the top-down pathway to semantic flow. They are complementary.
\subsection{Competitive Attention between Spatial and Semantic Flows}
The feature fusion method in FPN indiscriminately treats different information streams in a fixed feature aggregation way. Hence APN, shown in Figure  \ref{fig_4_1_2}, applies the proposed \textit{Competitive Attention (CA)} to suppress redundant features and to dynamically present emphasis between the semantic flow and spatial flow, implying an adaptive competition between them.

To implement \textit{CA}, we first aggregate the spatial information of feature maps of the spatial flow ${\mathbf{X}'}_l=[{\mathbf{x}'}_l^1,{\mathbf{x}'}_l^2,\ldots,{\mathbf{x}'}_l^C]$ and semantic flow $\mathbf{U}_{l}=[\mathbf{u}_{l}^1,\mathbf{u}_{l}^2,\ldots,\mathbf{u}_{l}^C]$, generating the global descriptors $\widehat{\mathbf{X}'}_l$ and $\hat{\mathbf{U}}_{l}$ for all channels of each flow:
\begin{equation}
\label{equ:se_gap_spa}
\widehat{\mathbf{x}'}_l^c=F_{sq}({\mathbf{x}'}_l^c)=\frac{1}{H_l \times W_l}\sum^{H_l}_{i=1}\sum^{W_l}_{j=1}[\mathbf{x'}_l^{c}]_{i,j},
\end{equation}
\begin{equation}
\label{equ:se_gap_sem}
\hat{\mathbf{u}}_{l}^c=F_{sq}(\mathbf{u}_{l}^c)=\frac{1}{H_{l} \times W_{l}}\sum^{H_{l}}_{i=1}\sum^{W_{l}}_{j=1}[\mathbf{u}_{l}^{c}]_{i,j},
\end{equation}
where $F_{sq}$ represents the squeeze operation to aggregate the information along the spatial dimension, $[\cdot]_{i,j}$ represents values of feature maps for the position (i, j), ${\mathbf{X}'}_l$, $\mathbf{U}_{l}$ $\in \mathbb{R}^{H_{l}\times W_{l} \times C}$, and $\widehat{\mathbf{X}'}_l, \hat{\mathbf{U}}_{l} \in \mathbb{R}^{1\times 1 \times C}$. The combination of $\widehat{\mathbf{X}'}_l$ and $\hat{\mathbf{U}}_{l}$ will be used as the joint input for the excitation operation to capture channel-wise complementary dependencies between spatial and semantic flows:
\begin{equation}
\label{equ:se_ex}
\begin{split}
\mathbf{S}_l&=F_{ex}([F_{sq}({\mathbf{X}'}_l);F_{sq}(\mathbf{U}_{l})];\mathbf{w}_{ex})\\
&=F_{ex}([\widehat{\mathbf{X}'}_{l};\hat{\mathbf{U}}_{l}];\mathbf{w}_{ex})\\
&=\sigma (ReLU([\widehat{\mathbf{X}'}_{l},\hat{\mathbf{U}}_{l}];\mathbf{w}_1);\mathbf{w}_2) \\
&=[\mathbf{s}_l^1,\mathbf{s}_l^2,\ldots,\mathbf{s}_l^C,\mathbf{s}_l^{C+1},\mathbf{s}_l^{C+2},\ldots,\mathbf{s}_l^{2C}],
\end{split}
\end{equation}
where $\sigma$ means the sigmoid activation, $[\cdot]$ refers to the concatenation along the channel dimension, parameters $\mathbf{w}_1\in \mathbb{R}^{\frac{2C}{t} \times 2C}$, $\mathbf{w}_2 \in \mathbb{R}^{2C \times \frac{2C}{t}}$, $F_{ex}$ denotes the excitation operation, and $\mathbf{S}_l \in \mathbb{R}^{1\times 1 \times 2C}$ is the result of $F_{ex}$ that will be divided into two parts to re-scale the weights of features ${\mathbf{X}'}_l$ and $\mathbf{U}_{l}$ respectively:
\begin{equation}\label{equ:se_scale_spa}
\begin{aligned}
\widetilde{\mathbf{X}'}_{l} &=F_{scale}(\mathbf{S}^{spa}_{l},{\mathbf{X}'}_l)\\
&=\mathbf{S}^{spa}_{l} \otimes {\mathbf{X}'}_l,
\end{aligned}
\end{equation}
\begin{equation}\label{equ:se_scale_sem}
\begin{aligned}
\tilde{\mathbf{U}}_{l} &=F_{scale}(\mathbf{S}^{sem}_{l},\mathbf{U}_{l})\\
&=\mathbf{S}^{sem}_{l}\otimes \mathbf{U}_{l},
\end{aligned}
\end{equation}
where $\otimes$ represents the element-wise multiplication, $F_{scale}$ denotes the scaling operation of CA, $\mathbf{S}^{spa}_{l} \in \mathbb{R}^{1\times 1 \times C}$ refers to $[\mathbf{s}_l^1,\mathbf{s}_l^2,\ldots,\mathbf{s}_l^C]$,and $\mathbf{S}^{sem}_{l} \in \mathbb{R}^{1\times 1 \times C}$ means $[\mathbf{s}_l^{C+1},\mathbf{s}_l^{C+2},\ldots,\mathbf{s}_l^{2C}]$. Finally, the overall feature fusion process with CA is reformulated as:
\begin{equation}\label{equ:se_scale}
\mathbf{P}'_l=F_{ca}^{spa}({\mathbf{X}'}_l,\mathbf{U}_{l})\otimes {\mathbf{X}'}_l+F_{ca}^{sem}({\mathbf{X}'}_l,\mathbf{U}_{l})\otimes \mathbf{U}_{l},
\end{equation}
where $F_{ca}^{spa}$ and $F_{ca}^{sem}$ refer to the modeling of CA for the spatial and semantic flow respectively.  Compared with Eq. \ref{equ:fpn_pl'}, Eq. \ref{equ:se_scale} adds two weights $\mathbf{S}^{spa}_l$ and $\mathbf{S}^{sem}_{l}$ for $\mathbf{X}'_l$ and $\mathbf{U}_{l}$, which is a softer feature aggregation form. Note that the difference between common attention mechanism such as SE and our method is that the range of modeling in our method is expanded into two flows, which simultaneously encodes the complementary relationship between spatial and semantic flows. Namely, our method can introduce the different information interactions into the attention modeling, instead of separately modeling each single information stream like the common attention mechanism.
\subsection{Spatial Collaborative Attention: trade-off between Global and Local}\label{sec:spa}
Competitive Attention, as the channel-wise attention, solely considers the global information for each feature map. However, for Chinese herbal recognition, it is necessary to take local clues into account and dynamically perform the trade-off between the global and local information. Moreover, from the spatial perspective, it is empirically confirmed that the semantic flow upsampled from the deeper layers usually contains the more global information, while the spatial flow is abundant in local details. If we consider both global and local cues of different levels, pixel-wise attention with the smaller granularity is desired. Thus we propose a \textit{Spatial Collaborative Attention} (SCA) that focuses on \textit{where} is an informative part in a feature map, shown in Figure  \ref{fig_4_1_2}, to model the collaboration and the supplement between the global and local information on different pyramid levels.

In order to achieve our goal, we squeeze features ${\mathbf{X}'}_l$ and $\mathbf{U}_{l}$ along the channel dimension and gain the descriptors of all channels for the position $(i,j)$:
\begin{equation}
\label{equ:sp_squeeze_lat}
[\check{\mathbf{X'}}_l]_{i,j}=G_{sq}([{\mathbf{X}'}_l]_{i,j})=\frac{1}{C}\sum^C_{c=1}{[{\mathbf{x'}}_l^{c}]_{i,j}},
\end{equation}
\begin{equation}
\label{equ:sp_squeeze_up}
[\check{\mathbf{U}}_{l}]_{i,j}=G_{sq}([\mathbf{U}_l]_{i,j})=\frac{1}{C}\sum^C_{c=1}{[\mathbf{u}_l^{c}]_{i,j}},
\end{equation}
where $G_{sq}$ refers to the squeeze operation of cross-channel global average pooling. Considering all positions of all feature maps, we obtain $\check{\textbf{X}'}_l \in \mathbb{R}^{H_l \times W_l \times 1}$ and $\check{\textbf{U}}_{l} \in \mathbb{R}^{H_l \times W_l \times 1}$, which will be concatenated to form $[{\check{\mathbf{X}'}}_l,\check{\mathbf{U}}_{l}]$ along the channel dimension and convolved by a standard convolution layer with the filter size of $3\times 3$ and the stride of 2. This process is denoted by the spatial squeeze operation $\hat{G}_{sq}$ and computed as:
\begin{equation}
\label{equ:sp_down}
\begin{aligned}
\varepsilon_l &=\hat{G}_{sq}([{\check{\mathbf{X}'}}_l,\check{\mathbf{U}}_{l}];\mathbf{w}_{(3\times 3)})\\
&=ReLU(f^{3\times 3}(\check{\mathbf{X}'}_l,\check{\mathbf{U}}_{l}])) =[\boldsymbol{\varepsilon}_l^1,\boldsymbol{\varepsilon}_l^2],
\end{aligned}
\end{equation}
where $\boldsymbol{\varepsilon}_l^1,\ \boldsymbol{\varepsilon}_l^2 \in \mathbb{R}^{\frac{H_l}{2}\times \frac{W_l}{2} \times 1}$. Subsequently, we obtain the spatial descriptor of all channels for each position and then take it as the input of the following excitation operation $G_{ex}$:
\begin{equation}
\label{equ:sp_ex}
\begin{aligned}
\boldsymbol{\xi}_l&=G_{ex}(F_{up}(\boldsymbol{\varepsilon}_l);\mathbf{w}_{(1\times 1)}) \\
&=G_{ex}(\mathbf{e}_l;\mathbf{w}_{(1\times 1)}) =\sigma(f^{1\times 1}(\mathbf{e}_l)) =[\boldsymbol{\xi}_l^1, \boldsymbol{\xi}_l^2],
\end{aligned}
\end{equation}
where $ \mathbf{e}_l = F_{up}(\varepsilon_l) $. As a result, we obtain two attention masks $\boldsymbol{\xi}_l^1,\ \boldsymbol{\xi}_l^2 \in \mathbb{R}^{H_l \times W_l \times 1}$ to re-scale two flows on the pixel-level. When combining CA with SCA, we first perform the tensor multiply for the activation of CA and SCA respectively:
\begin{equation}
\label{equ:comb_spa}
\textbf{M}_l^{spa}=\boldsymbol{\xi}_l^1 \otimes {\mathbf{S}}_l^{spa},
\end{equation}
\begin{equation}
\label{equ:comb_sem}
\textbf{M}_l^{sem}=\boldsymbol{\xi}_l^2 \otimes {\mathbf{S}}_l^{sem},
\end{equation}
where $\textbf{M}_l^{spa},\  \textbf{M}_l^{sem} \in \mathbb{R}^{H_l \times W_l \times C}$. 

Furthermore, we also deploy $1\times 1$ convolution layers after multiplying since these two attention modeling processes are not mutually independent, and then we obtain $\hat{\textbf{M}}_l^{spa},\ \hat{\textbf{M}}_l^{sem} \in \mathbb{R}^{H_l \times W_l \times C}$.
Compared with Eq. \ref{equ:se_scale_spa} and Eq. \ref{equ:se_scale_sem}, $\hat{\textbf{M}}_l^{spa}$ and $  \hat{\textbf{M}}_l^{sem}$ are composed of two parts: ${\mathbf{S}}^{spa(sem)}_{l} \in \mathbb{R}^{1\times 1 \times C}$ and $\boldsymbol{\xi}_l^{1(2)} \in \mathbb{R}^{H_l \times W_l \times 1}$.  The overall attention combining CA with SCA is denoted by CSCA module and its more details are shown in Figure  \ref{fig_4_1_2} (right). As a result, the final fused features of the level $l$ can be gained by scaling $\mathbf{X}'_l$ and $\mathbf{U}_{l}$ with $\hat{\textbf{M}}_l^{spa}$ and $  \hat{\textbf{M}}_l^{sem}$:
\begin{equation}
\label{equ:comb_pl}
\mathbf{P}'_l=\hat{\textbf{M}}_l^{spa} \otimes \mathbf{X}'_l + \hat{\textbf{M}}_l^{sem} \otimes \mathbf{U}_{l}
\end{equation}
Compared with Eq. \ref{equ:se_scale}, Eq. \ref{equ:comb_pl} considers both the global and local information.

\subsection{Explore Parametric Spatial Collaborative Attention}
\begin{figure}
\centering
\includegraphics[scale=0.4]{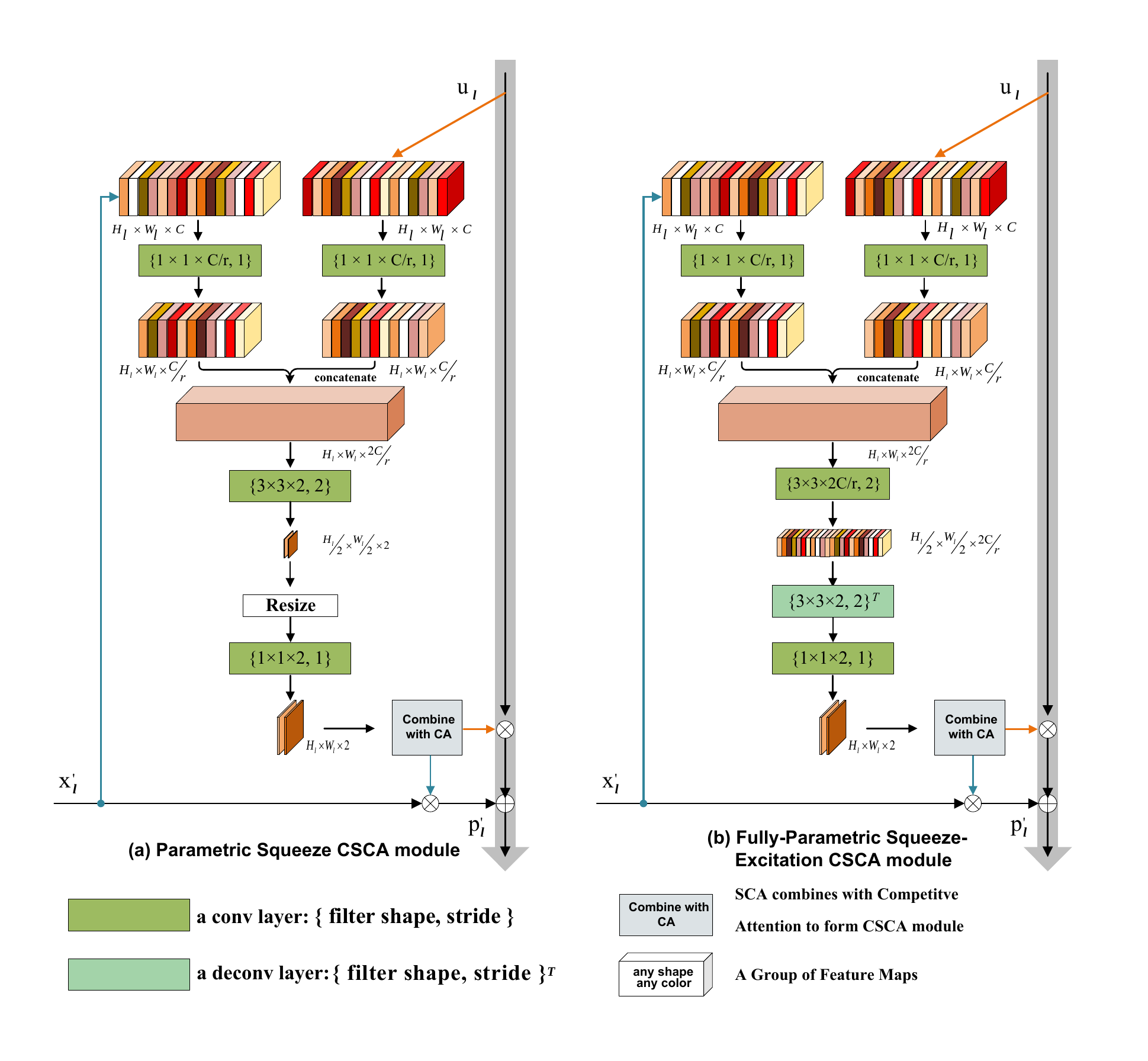}
\caption{Parameteric Squeeze CSCA module (left) and Fully-Parametric Squeeze-Excitation CSCA module (right). Their combination with Competitive Attention is not shown for brevity.}
\label{fig_4_4}
\end{figure}
Let SCA-$\alpha$ denote the parameter-free SCA as described in Sec.\ref{sec:spa}, it can offer an effective mechanism to trade off the global and local information but ignore their complementary relations between various channels. To solve the issue, we introduce parameters into the squeeze and excitation operation.\\
\textbf{Parametric Squeeze Operator.} Firstly, given the features ${\mathbf{X}'}_l$, $\mathbf{U}_{l} \in \mathbb{R}^{H_l \times W_l \times C}$ as the input, we define the parametric \textit{channel-squeeze} operator $G^\text{+}_{sq}(\cdot)$ to reduce the parameter overhead:
\begin{equation}
\label{equ:soft_1_sq_spa}
\check{\mathbf{X}'}_l^\text{+} =G^\text{+}_{sq}(\mathbf{X}'_l;\mathbf{w}_{(1\times 1)})=f^{1\times 1}(\mathbf{X}'_l)
\end{equation}
\begin{equation}
\label{equ:soft_1_sq_sem}
\check{\mathbf{U}}_{l}^\text{+}=G^\text{+}_{sq}(\mathbf{U}_{l};\mathbf{w}_{(1\times 1)})=f^{1\times 1}(\mathbf{U}_{l})
\end{equation}
where $\check{\mathbf{X}'}_l^\text{+},\  \check{\mathbf{U}}_{l}^\text{+} \in \mathbb{R}^{H_l \times W_l \times \frac{C}{r}}$, and $\mathbf{w}_{(1\times 1)}$ means the weight of a $1\times 1$ convolution layer.
Compared with the parameter-free squeeze operator $G_{sq}$ in Eq. \ref{equ:sp_squeeze_lat} and \ref{equ:sp_squeeze_up}, features $\mathbf{X}'_l,\ \mathbf{U}_{l} \in \mathbb{R}^{H_l \times W_l \times C}$ are transformed into a lower-dimension space with the reduction ratio $r$ and then produce $\check{\mathbf{X}'}_l^\text{+}$, $\check{\mathbf{U}}_{l}^\text{+} \in \mathbb{R}^{H_l \times W_l \times \frac{C}{r}}$, instead of global signals of all channels $\check{\textbf{X}'}_l$, $\check{\textbf{U}}_{l} \in \mathbb{R}^{H_l \times W_l \times 1}$.

Feature maps of different channels are not mutually independent~\cite{wu2018group} and usually focus on different spatial regions. As a consequence, it is reasonable to believe that feature maps of multiple channels participating in the modelling process is beneficial to making a trade-off between the global and local information. Here the modified SCA and CSCA module with parametric squeeze operators are denoted respectively as \textit{SCA-}$\theta$ and \textit{CSCA-}$\theta$, whose details are shown in Figure  \ref{fig_4_4} (left).
\\
\textbf{Fully-Parametric Squeeze-Excitation Operator.} As discussed above, we only parameterized the squeeze operation while the excitation operation is not considered. Now we extend the parameterization range into the excitation operator and propose \textit{Fully-Parameteric Squeeze-Excitation SCA} and \textit{CSCA}, denoted by \textit{SCA-}$\theta^\text{+}$ and \textit{CSCA-}$\theta^\text{+}$, as shown in Figure  \ref{fig_4_4} (right). We first change the number of filters in the \textit{spatial squeeze operator} $\hat{G}_{sq} $ in Eq. \ref{equ:sp_down} into $ \hat{G}_{sq}^\text{+}$:
\begin{equation}
\label{equ:soft2_down}
\begin{aligned}
\boldsymbol{\varepsilon}_l^\text{+} &=\hat{G}_{sq}^\text{+}([{\check{\mathbf{x}'}}_l^\text{+},\check{\mathbf{u}}_{l}^\text{+}];\mathbf{w}_{(3\times 3)})\\
&=ReLU(f^{3\times 3}([{\check{\mathbf{X}'}}_l^\text{+},\check{\mathbf{U}}_{l}^\text{+}])),
\end{aligned}
\end{equation}
where $\boldsymbol{\varepsilon}_l^\text{+} \in \mathbb{R}^{\frac{H_l}{2} \times \frac{W_l}{2} \times \frac{2C}{r}}$ is used as the input of the parametric excitation operation implemented by deconvolution $f^{(3\times 3)^T}$:
\begin{equation}
\label{equ:soft_2_up}
\begin{aligned}
\boldsymbol{\xi}_l^\text{+}&=G_{ex}^\text{+}(f^{(3\times 3)^T}(\boldsymbol{\varepsilon}_l^{\text{+}});\mathbf{w}_{(1\times 1)}) \\
&=G_{ex}^\text{+}(\mathbf{e}_l^+;\mathbf{w}_{(1\times 1)}) =f^{1\times 1}(\mathbf{e}_l^+) =[{\boldsymbol{\xi}^\text{+}}_l^1, {\boldsymbol{\xi}^\text{+}}_l^2],
\end{aligned}
\end{equation}
where $\mathbf{e}_l^+ \in \mathbb{R}^{H_l \times W_l \times 2}$ and ${\boldsymbol{\xi}^\text{+}}_l^1, {\boldsymbol{\xi}^\text{+}}_l^2 \in \mathbb{R}^{H_l \times W_l \times 1}$. Compared with Eq. \ref{equ:sp_ex}, $G_{ex}^\text{+}$ is learnable and considers the relations between channels.

In summary, the parametric Squeeze-Excitation operation provides a larger modelling space for the collaborative features fusion. Although their parameter sizes slightly increase, they still satisfy the requirement of lightweight for the attention mechanism.
\subsection{Herbal Recognition Framework}
APN can be applicable to general multi-granularity recognition tasks such as Chinese herbal recognition, where CSCA is used to dynamically select features with different scales in demand for the recognition granularity. CSCA is composed of CA module and SCA module, where both CA and SCA are applied to the semantic and spatial flows. Their inputs are the combined features from both semantic flows and spatial flows, so that there are the competition and collaboration between two flows in the process of attentional modeling.
In order to design the herbal recognition framework, APN takes pre-act ResNet~\cite{he2016identity} and VGG~\cite{simonyan2015very} as the backbone networks. Different from FPN predicting on each level for object detection, the output $\mathbf{P}_l \in \mathbb{R}^{H_l\times W_l \times C}$ of each level on top-down pathway in our methods will be globally pooled to form  $\hat{\mathbf{P}}_l \in \mathbb{R}^{1\times 1 \times C}$, which are then concatenated to form $\hat{\mathbf{P}}=[\hat{\mathbf{P}}_1,\hat{\mathbf{P}}
_2,\ldots,\hat{\mathbf{P}}_L]$. Finally, $\hat{\mathbf{P}}$ is sequentially forwarded to a fully-connected layer and softmax classifier to perform the classification.
\section{Chinese-Herbs Dataset} \label{sec:dataset}
\begin{figure*}
\centering
\includegraphics[scale=0.49]{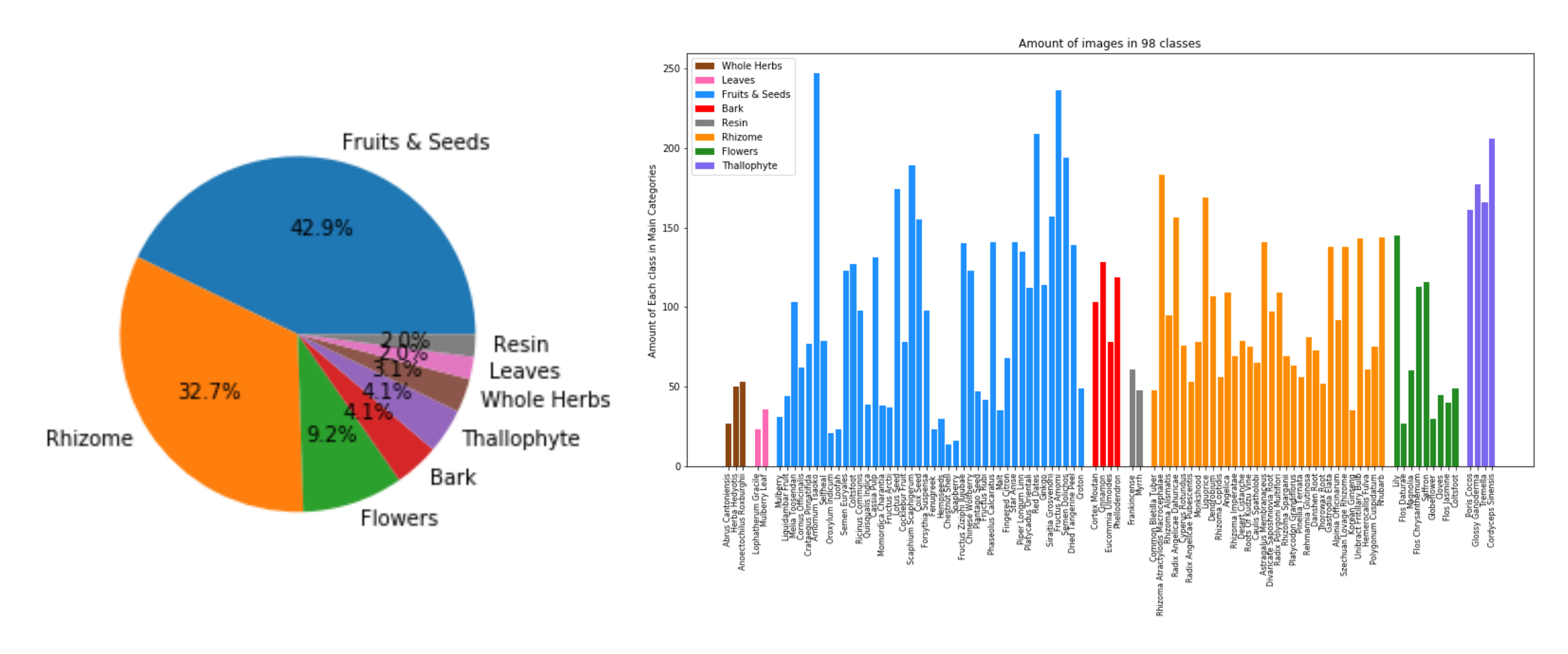}
\caption{Distribution of Chinese Herbs Species (left) and the number of images for each classes in CNH-98 (right).}
\label{fig_pie_bar}
\end{figure*}
As there are no standard Chinese-Herbs Datasets available at present, we construct a novel Chinese-Herbs Dataset(CNH-98) having 9184 images with 98 classes, which mainly consists of natural plants and thallophyte, covering the common Chinese herbs. Subsequently it is divided randomly into training and validation sets with a proportion of 4:1. Figure  \ref{fig_exhibition} shows some examples of CNH-98. The datasets can be available\footnote{\url{https://github.com/scut-aitcm/Chinese-Herbs-Dataset}}. \\
\begin{table}
\center
\caption{Main species of CNH-98 dataset and their corresponding examples.}
\scriptsize
\begin{tabular}{l c}
\hline
\textbf{Main Species}&\multirow{1}{*}{\textbf{Herbs Examples}}\\
\hline
\multirow{3}{4cm}{Fruits \& Seeds} & Star Anise, Siraitia Grosvenorii,\\
~&Ginkgo, Chinese Wolfberry,\\ ~&SElfheal, Fructus Arctii, etc.\\
\hline
\multirow{3}*{Rhizome} & {Liquorice, Thorowax Root,}\\
~&Rhizoma Alismatis, \\
~&Unibract Fritillary Bulb, etc.\\
\hline
\multirow{3}*{Flowers} &{Saffron, Flos Daturae,}\\
~&{Cloves, Magnolia, Coltsfoot,}\\
~&{Flos Jasmine, Lily, etc.}\\
\hline
\multirow{2}*{Bark} &{Cinnamon, Cortex Moutan,}\\
~&{Eucommia Ulmoides, etc.}\\
\hline
\multirow{2}*{Thallophyte} &{Glossy Ganoderma, Tremella
,}\\
~&{Cordyceps Sinensis, etc.}\\
\hline
\multirow{2}*{Whole Herbs} &{Abrus cantoniensis,}\\
~&{Anoectochilus roxburghii, etc.}\\
\hline
Leaves&{Lophatherum Gracile, etc.}\\
\hline
Resin&{Frankincense, Myrrh, etc.}\\
\hline
\end{tabular}
\label{tab_herbs_details}
\end{table}
\subsection{Dataset Collection}
In this dataset, about 80\% of the images are collected from the Google images~\cite{GoogleImage}. Moreover, the others are acquired by taking photos ourselves in the medicinal herbs stores. The number of images per class in CNH-98 ranges 14 to 246, where there are more than 41 classes having over 100 images. The size of images in CNH-98 covers 150-1500 pixels. Considering the quality of images and annotations, all images are filtered and manually checked by multi-human annotators strategy. Thus the quality of data is reliable. The detailed collection process is as follows:
\\
\textit{\textbf{Constructing Candidate Images.}} First of all, it is necessary to determine the candidate list of Chinese herbs classes before collecting data. There are a great variety of Chinese herbs, the most of which are extremely rare. Since the rare herbs are difficult to collect, we need to select the common classes for constructing candidate list of herbs classes. We thus crawl three standard Chinese herbs websites for the lists of all classes of herbs, and then we have take the intersection of classes list from three websites as candidate list of herbs classes. Based on the class candidate list, candidate images of Chinese herbs for each class are collected from Google images~\cite{GoogleImage}. Images from Google provides the label information, which simplified labeling. For each class of candidate list, about 300 images are downloaded as the part of candidate images. The rest candidate images are hand-collected by taking photos of herbs in the medicine herbs stores, which accounted for 20\%.
\\
\textit{\textbf{Quality Control on Images.}} To ensure the diversity and the quality of samples, the quality control on images primarily consists of two aspects: filtering duplicate and inferior images. For the former, the perceptual hash algorithm~\cite{zauner2010implementation} is used to remove the duplicate images from the candidate images. For the latter, images are removed if any of the following conditions is satisfied: (i) The image contains no herbs; (ii) The proportion of herbs in image is too small; (iii) The image is gray.Furthermore, both duplicate images and inferior images to be removed are checked by two annotators, where the third annotators would join checking when disagreement occurring.
\\
\textit{\textbf{Multi-human Annotations.}} Each hand-collected image by taking photos is required for hand-annotations, which is labeled by two annotators independently. When there are different labels for the same image, the third annotator would participate in the annotation. If there are two annotators labeling it as the same class, the image would be annotated by the corresponding label. If not, the image would be considered as a confused image and then dropped out.

Finally, in order to ensure the quantity of each class, the classes of the top 90\% are reserved according to the number of images. Eventually we obtain the database that contains 98 classes. The complete class list is presented in the supplementary material.
\subsection{Grouping of Chinese Herbs}
The classes of CNH-98 dataset cover the natural plants and thallophyte, and herbs are always obtained from plant organs, such as fruit and rhizome. Human recognize these herbs not only by the texture, but also by the shape and color, indicating that there is a great visual similarity between the herbs from the same plant organs. If we define a group of herbs from the same plant organ as a species and the herbs of thallophyte as another species. Herbs belonging to different species are easier to classify while those belonging to the same species are so similar that they should be classified with more fine-grained cues. The classes and species of herbs are similar to that of other database that contains people, cars, and buildings as classes, having the semantic meaning that can be utilized to improve the recognition performance.
In CNH-98, there are eight species, including Fruits \& Seeds, Rhizome, Flowers, Bark, Thallphyte, Whole Herbs, Leaves, Resin. Their examples are shown in Figure \ref{fig_exhibition} and Table  \ref{tab_herbs_details}.

It can be observed that the herbs of different species are easily distinguished, whereas the herbs of the same species are similar with the difference only in details. This exactly supports for the motivation of our proposed methods.
At the same time, Figure  \ref{fig_pie_bar} (left) presents the distribution of the number of Chinese herbs classes for eight species, where a majority of classes are Fruits \& Seeds and Rhizome, including 42 and 32 classes respectively. It can be seen from Figure  \ref{fig_pie_bar} (right) that CNH-98 dataset is relatively unbalanced.

\section{Experiments}
\label{sec:experiments}
Lots of experiments are conducted on our proposed dataset CNH-98 to validate the proposed method, where recently proposed methods are compared.
\subsection{Experimental Setup} \label{sec:implem}

\begin{table*}[t]
\centering
\scriptsize
\caption{\textbf{Structure of Attentional Pyramid Networks} with pre-act ResNet-18 and ResNet-34 as backbone on the bottom-up pathway and CSCA modules on the top-down pathway. The arrow ($\downarrow / \uparrow$) refers to the propagation direction of feature information steams. \textit{c-avgpool} on the CSCA modules means cross-channel average pooling. @3 refers that there are 3 duplicated CSCA modules in total, corresponding to 3 output sizes respectively.}
\begin{tabular}{c c c c c c}
\hline
\multirow{3}{*}{\textbf{Output Size}} & \multicolumn{2}{c}{\textbf{Bottom-up} ($\downarrow$)}  & \multicolumn{3}{c}{\textbf{Top-down} ($\uparrow$)}\\
\cline{2-6}
~&\multirow{2}{*}{\textbf{18-layer}} & \multirow{2}{*}{\textbf{34-layer}} & \multicolumn{2}{c}{\textbf{Input Size}}  & \multirow{2}{*}{\textbf{CSCA Module}}\\
\cline{4-5}
~&~&~& \textbf{Lateral} & \textbf{Upsampling} & ~\\
\hline
$112\times 112$ & \multicolumn{2}{c}{$7\times 7$, 64, stride 2} & - & - & -\\
\hline
\multirow{2}{*}{$56\times 56$} & \multicolumn{2}{c}{$3\times 3$ maxpool, stride 2} & - & - & -\\
\cline{2-6}
~&
$
\begin{bmatrix}
\begin{bmatrix}
3 \times 3 , 64\\
3 \times 3 , 64
\end{bmatrix} \times 2 \\
1 \times 1 , 256
\end{bmatrix}
$
&
$
\begin{bmatrix}
\begin{bmatrix}
3 \times 3 , 64\\
3 \times 3 , 64
\end{bmatrix} \times 3 \\
1 \times 1 , 256
\end{bmatrix}
$
& $56\times 56$ & $28\times 28$ &
\multirow{3}{*}
{
$
\begin{bmatrix}
\text{upsampling}\\
\begin{bmatrix}
\begin{bmatrix}
\text{avgpool}\times 2\\
fc, [32,512]\\
\end{bmatrix}\\
\begin{bmatrix}
\text{c-avgpool}\times 2\\
3\times 3,2\\
1\times 1,2\\
\end{bmatrix}\\
\begin{bmatrix}
1\times 1, 256
\end{bmatrix}\times 2\\
\end{bmatrix}\\
3\times 3, 256
\end{bmatrix} @3
$
}\\
\cline{1-5}
$28\times 28$ &
$
\begin{bmatrix}
\begin{bmatrix}
3 \times 3 , 128\\
3 \times 3 , 128
\end{bmatrix} \times 2 \\
1 \times 1 , 256
\end{bmatrix}
$
&
$
\begin{bmatrix}
\begin{bmatrix}
3 \times 3 , 128\\
3 \times 3 , 128
\end{bmatrix} \times 4 \\
1 \times 1 , 256
\end{bmatrix}
$
& $28\times 28$ & $14\times 14$ &~\\
\cline{1-5}
$14\times 14$ &
$
\begin{bmatrix}
\begin{bmatrix}
3 \times 3 , 256\\
3 \times 3 , 256
\end{bmatrix} \times 2 \\
1 \times 1 , 256
\end{bmatrix}
$
&
$
\begin{bmatrix}
\begin{bmatrix}
3 \times 3 , 256\\
3 \times 3 , 256
\end{bmatrix} \times 6 \\
1 \times 1 , 256
\end{bmatrix}
$
& $14\times 14$ & $7\times 7$ &~\\
\hline
$7\times 7$&
$
\begin{bmatrix}
\begin{bmatrix}
3 \times 3 , 512\\
3 \times 3 , 512
\end{bmatrix} \times 2 \\
1 \times 1 , 256
\end{bmatrix}
$
&
$
\begin{bmatrix}
\begin{bmatrix}
3 \times 3 , 512\\
3 \times 3 , 512
\end{bmatrix} \times 3 \\
1 \times 1 , 256
\end{bmatrix}
$
&
$7\times 7$ & - & -\\
\hline
$1\times 1$ & \multicolumn{5}{c}{$\text{global average pool}\times 4$, 98-d fc, softmax}\\
\hline
\end{tabular}
\label{tab_details of networks}
\end{table*}

In order to make the fair comparison, APN with CA, SCA, CSCA and their parameterized variants are trained with the same optimization schemes as that of FPN. It should be emphasized that all hyper-parameters keep consistent for all compared models except for the channel dimension reduction ratio $r$ of squeeze operators in the experiments on Parametric Spatial Collaborative Attention. Table \ref{tab_details of networks} lists the detailed architectures of APN-CSCA with pre-act ResNet-18 and ResNet-34 as the backbone. Compared with the typical pre-act ResNet, residual blocks of our backbone network on the bottom-up pathway have an additional $1\times 1$ convolution as a lateral connection structure, where the default channel dimension is 256.

In order to ensure the best performance of the compared methods, we follow the practice in~\cite{hu2019squeeze} where the scale of the input image is consistent with the one in our proposed dataset. Namely, all methods are trained with the standard data augmentation: the translation/mirroring is adopted and the $224\times 224$ crop is randomly sampled. All images are normalized with mean values and standard deviations. When testing, the test sample is firstly resized to the size of $256\times 256$ and then obtain the $224\times 224$ central crop. All models are trained by the optimizer SGD with 0.9 Nesterov momentum from scratch. The upsampling strategy is achieved by the bilinear interpolation elsewhere specified.
All models are trained with batch size 64 and 300 epochs. We choose the best validation score of 300 epochs as the result of each run and carry on the 5-fold cross-validation. The learning rate is initialized to 0.1 and divided by 5 at epoch 120, 200, 260. Weight decay is adopted with 0.0005. 

To assess the performance, several commonly used metrics are used, including top-1 accuracy, macro-precision (P), macro-recall (R) and macro-F1 score (F1). For analysis of complexity, we also report the parameter sizes and FLOPs (the number of floating-point operations) of each model, following the practice in~\cite{he2016identity}.

\subsection{Ablation Studies} \label{sec:ablation}
\begin{table*}[t]
\centering
\scriptsize
\caption{Evaluation results of models with different components on datasets CNH-98. The best records of comparative groups are \textbf{bold} and the best records of all models with the same depth are \textbf{bold} and {\color{red}red}. The variants of our methods are also shown in \textbf{bold}.}
\begin{tabular}{c l c c c c c c c c c c}
\toprule
\multirow{2}{*}{\textbf{Backbone}}&\multirow{2}{*}{\textbf{Methods}}&\multirow{2}{*}{\textbf{CA}}&\multicolumn{3}{c}{\textbf{SCA}} &\multirow{2}{*}{\textbf{params}}&\textbf{FLOPs}&\multirow{2}{*}{\textbf{top-1 (\%)}}&\multirow{2}{*}{\textbf{P}}&\multirow{2}{*}{\textbf{R}}&\multirow{2}{*}{\textbf{F1}}\\
\cline{4-6}
~&~&~ &\textbf{$\alpha$}&\textbf{$\theta$}&\textbf{$\theta^\text{+}$}&~&($\times 10^9$)\\
\hline
\multirow{7}{*}{Res-18~\cite{he2016identity}} & (a) FPN~\cite{lin2017feature} &~&~&~&~&13.3M&4.34&91.9&89.4&88.5&89.0\\
\cline{2-12}
~&\textbf{(b) APN-CA} & \checkmark &~&~&~&13.4M&4.34&92.9&90.7&90.8&90.7\\
~&\textbf{(c) APN-SCA} & ~&\checkmark &~&~&13.3M&4.34&92.5&90.9&90.0&90.5\\
~&\textbf{(d) APN-CSCA} &\checkmark & \checkmark & ~&~ & 13.8M&4.88&\textbf{93.5}&\textbf{91.7}&\textbf{91.2}&\textbf{91.4}\\
\cline{2-12}
~&\textbf{(e) APN-CSCA-$\theta$} &\checkmark & ~ & \checkmark &~ & 13.8M&4.90&93.5&92.2&91.7&91.9\\
~&\textbf{(f) APN-CSCA-$\theta^{\text{+}}$} &\checkmark & ~ & ~ & \checkmark & 13.8M&4.90&\textbf{\color{red}93.7}&\textbf{\color{red}92.7}&\textbf{\color{red}92.2}&\textbf{\color{red}92.4}\\
\hline

\multirow{6}{*}{Res-34~\cite{he2016identity}} &(a) FPN~\cite{lin2017feature} &~&~&~&~&23.4M&6.19&92.3&91.0&90.3&90.7\\
\cline{2-12}
~&\textbf{(b) APN-CA} & \checkmark &~&~&~&23.5M&6.19&93.5&90.8&91.1&90.9\\
~&\textbf{(c) APN-SCA} & ~&\checkmark &~&~&23.4M&6.19&92.7&90.7&91.0&90.8\\
~&\textbf{(d) APN-CSCA} &\checkmark & \checkmark & ~&~ & 23.9M&6.73&\textbf{93.8}&\textbf{92.2}&\textbf{91.6}&\textbf{91.9}\\
\cline{2-12}
~&\textbf{(e) APN-CSCA-$\theta$} &\checkmark & ~ & \checkmark &~ & 23.9M&6.75&93.9&92.7&92.3&92.5\\
~&\textbf{(f) APN-CSCA-$\theta^{\text{+}}$} &\checkmark & ~ & ~ & \checkmark & 23.9M&6.75&\textbf{\color{red}94.0}&\textbf{\color{red}92.9}&\textbf{\color{red}92.7}&\textbf{\color{red}92.8}\\
\bottomrule
\end{tabular}
\label{tab:fair_cmp}
\end{table*}
In this section, the ablation experiments are conducted on CNH-98 to validate the effectiveness of all components of APN, including CA, parameter-free SCA-$\alpha$, CSCA, and their corresponding parametric version (SCA-$\theta$, SCA-$\theta^\text{+}$, CSCA-$\theta$ and CSCA-$\theta^\text{+}$), where FPN is taken as the baseline. The experimental results are presented in Table. \ref{tab:fair_cmp}, where methods (b)-(d) are free parametric approaches. It can be observed that in most cases both APN-CA and APN-SCA outperform the baselines without too many extra parameters and FLOPs, indicating that both CA and SCA are effective. When combining CA with SCA (i.e. CSCA), our method achieves further significant improvements for all metrics, indicating that CA and SCA are complementary, although with a few of increases of FLOPs. When jointly modelling, CA and SCA can obtain the feature information more completely from two different perspectives so that they could reduce the possibility of falling into a locally optimal point. Specifically, compared with FPN-34, APN-CSCA-18 is even preferable for all metrics, but its size of parameters significantly reduces to about 58\% of FPN-34 and FLOPs reduce to about 78\% of FPN-34. This reveals that our methods can obtain the better performance with fewer parameters and FLOPs since they benefit from dynamically adjusting weights of features with different scales and then lead to the more efficient features fusion.

\begin{figure*}
\centering
\includegraphics[scale=0.6]{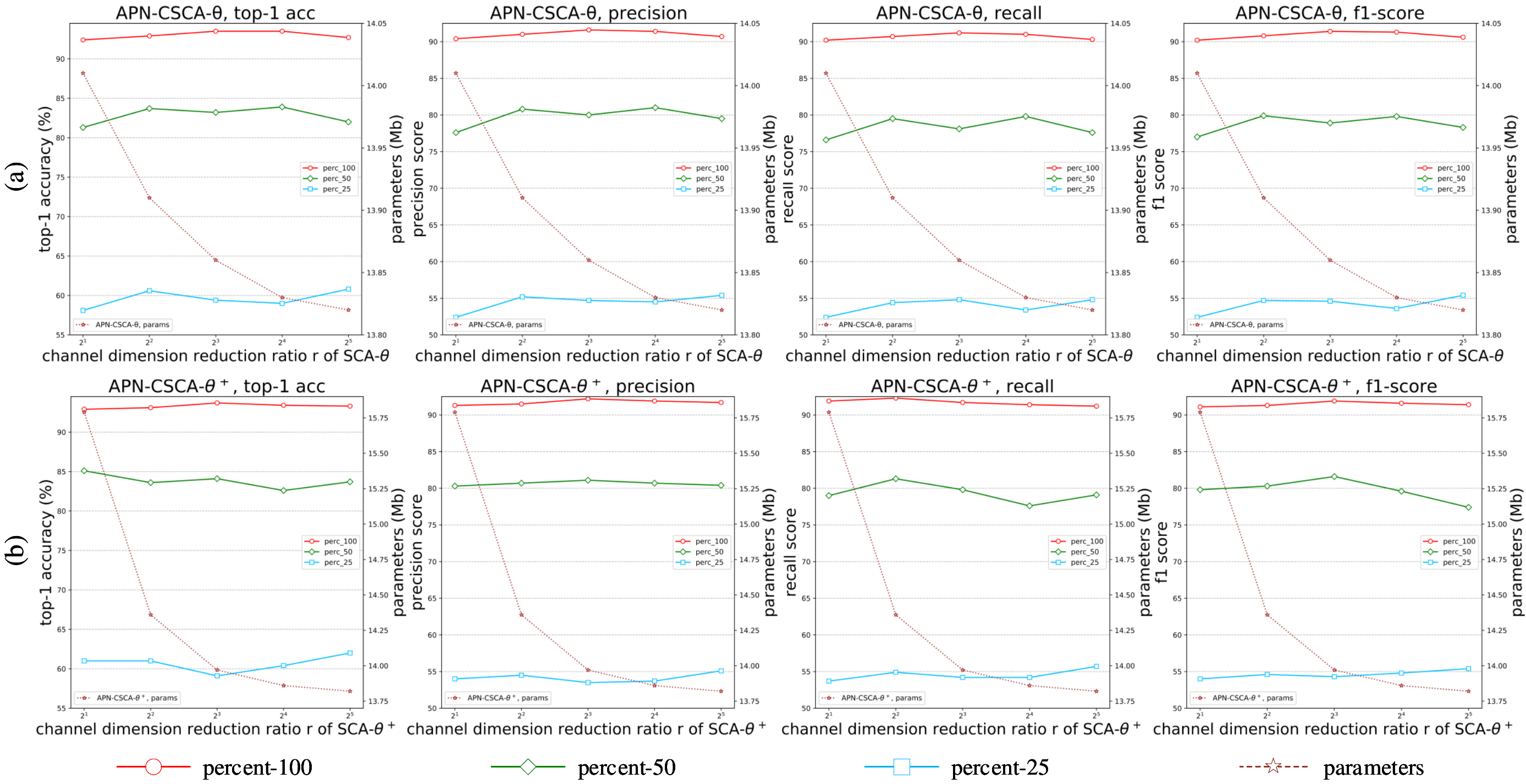}
\caption{Comparisons with different reduction ratio $r$ of SCA-$\theta$/$\theta^+$ in APN-CSCA-$\theta$/$\theta^+$-18. Here, the reduction ratio $t$ of CA is set by 16.}
\label{fig:ratio}
\end{figure*}

On the other hand, Table \ref{tab:fair_cmp} (e) and (f) present the results of the parametric spatial attention. It can be found that APN-CSCA-$\theta$ with parametric squeeze operators can bring additional improvements. Moreover, APN-CSCA-$\theta^\text{+}$ reaches a higher peak. These results suggest that the parametric SCA can make more contributions to the improvement of classification. This is because the parametric modeling makes it more possible to consider the relationship between different channels, resulting in the more efficient features fusion.\\
\textit{\textbf{Analyzing the reduction ratio of CSCA.}} Both the reduction ratio $t$ in CA and $r$ in parametric SCA are hyper-parameters, allowing us to vary the capacity and computational cost of CSCA in the network. In order to investigate the trade-off between the performance and computational cost determined by the reduction ratio $t$ or $r$, a series of experiments with APN-CSCA-$\theta$/$\theta^+$-18 for a range of different ratio t/r are conducted, following the settings in ~\cite{hu2019squeeze}. On the one hand, the results of the reduction ratio $t$ in CA on CNH-98 are reported in Table \ref{tab:ca_t}, which shows that the performance cannot be improved monotonically if only increasing t. When t = 16, our model achieves the good balance between the performance and complexity. On the other hand, considering that the parameters of model are closely related to the size of dataset, experiments are conducted for ratio $r$ in SCA-$\theta$/$\theta^+$ on a series of datasets that are randomly sampled 25\%, 50\% and 100\% of CNH-98. It can be seen from Figure  \ref{fig:ratio} that performances on these datasets are robust to a wide range of reduction ratios r. Furthermore, the highest performance is obtained at the smaller reduction ratio r (refers to more parameters) when the data becomes larger. Specifically, our methods reach the good balance at r=8.
\begin{table}[t]
    \centering
    \tiny
    \caption{Comparisons with different reduction ratio $t$ of CA in APN-CSCA-$\theta$/$\theta^+$-18. Here, the reduction ratio $r$ of SCA-$\theta$/$\theta^+$ is set by 8. The best results are \textbf{bold}.}
    \begin{tabular}{c c c c c c c}
    \toprule
         \textbf{Methods} &\textbf{Ratio} $t$& \textbf{Params} & \textbf{top-1(\%)} & \textbf{P} & \textbf{R} & \textbf{F1} \\
         \hline
         \multirow{5}{*}{APN-CSCA-$\theta$} & $2^1$ & 14.5M & 91.9 & 90.6 & 90.4 & 90.5\\
         ~& $2^2$ & 14.2M & 92.3 & 90.7 & 89.5 & 89.9\\
         ~& $2^3$ & 13.9M & 92.5 & 91.0 & 90.2 & 90.5 \\
         ~& $2^4$ & 13.8M & \textbf{93.5} & \textbf{92.2} & \textbf{91.7} & \textbf{91.9} \\
         ~& $2^5$ & 13.8M & 92.7 & 91.3 & 90.7 & 90.9 \\
         \hline
         \multirow{5}{*}{APN-CSCA-$\theta^+$} & $2^1$ & 14.7M & 91.9 & 89.8 & 89.7 & 89.7\\
         ~& $2^2$ & 14.3M & 92.4 & 90.9 & 90.4 & 90.7\\
         ~& $2^3$ & 14.1M & 92.9 & 91.1 & 91.2 & 91.1 \\
         ~& $2^4$ & 13.8M & \textbf{93.7} & \textbf{92.7} & \textbf{92.2} & \textbf{92.4} \\
         ~& $2^5$ & 13.8M & 93.1 & 91.7 & 91.5 & 91.4 \\
         \bottomrule
    \end{tabular}

    \label{tab:ca_t}
\end{table}
\begin{table*}[t]
\center
\caption{Results with VGG as backbone. Our methods are shown in \textbf{bold}.}
\vspace{1.5mm}
\begin{tabular}{l l c c c c c c c}
\toprule
\textbf{Backbone}&\textbf{Model} & \textbf{Size} & \textbf{Params} & \textbf{FLOPs} & \textbf{top-1(\%)} & \textbf{P}&\textbf{R}&\textbf{F1}\\
\midrule
\multirow{4}{*}{VGG~\cite{simonyan2015very}} & FPN-VGG~\cite{lin2017feature} & 11 & 11.5M &10.11$\times 10^9$ & 91.9&89.5&89.6&89.5\\
\cline{2-9}
~&\textbf{APN-CSCA}& 11 & 11.9M &10.65$\times 10^9$ & 93.4&91.7&92.0&91.9\\
~&\textbf{APN-CSCA-$\theta$}& 11 &12.0M&10.72$\times 10^9$& 93.5 & 91.9&\textbf{92.2} & 92.1\\
~&\textbf{APN-CSCA-$\theta^+$}& 11 &12.1M &10.75$\times 10^9$& \textbf{93.9} & \textbf{92.2}&92.1&\textbf{92.2}\\
\bottomrule
\end{tabular}
\label{tab:wrn}
\end{table*}
\\
\textbf{Comparison with different backbone networks.} In order to validate the independence of the used backbone networks in APN, the experiments are conducted on CNH-98, where VGG is taken as the backbone networks. Compared with FPN, it can be seen from the experimental results, shown in Table \ref{tab:wrn}, that both APN-CSCA and its parametric version bring apparent improvements of the performance. For example, there are improvements by 2.0\% for APN-CSCA-$\theta^+$ in terms of top-1. This indicates that our method can choose the better backbone network when applied to solve the practical problems.

\subsection{Comparison with recently proposed methods} \label{sec:cmp_soa}
\begin{table}[t]
\center
\tiny
\caption{Comparison with recently proposed methods on CNH-98.}
\begin{tabular}{l c c c c c c}
\toprule
\multirow{1}{*}{\textbf{Model}} & \textbf{Depth} & \textbf{Params} & \textbf{top-1(\%)} & \textbf{P} & \textbf{R} & \textbf{F1}\\
\hline
Network in Network~\cite{lin2014network} & - & 2.3M & 80.1 & 76.4 & 77.3&76.8
\\
VGG-11~\cite{simonyan2015very} & 11 & 129.1M & 88.1&84.3&83.8&84.1
\\
VGG-19~\cite{simonyan2015very} & 19 & 140.0M & 85.8&82.1&81.7&81.9
\\
GoogleNet~\cite{ioffe2015batch} & 22 & 7.0M & 92.0&89.7&89.5&89.6
\\
pre-act ResNet-18~\cite{he2016identity} & 18 & 11.7M  &91.7& 89.1 & 88.1 & 88.6\\

WRN-18-1.5~\cite{zagoruyko2016wide} & 18 & 25.2M & 92.7&\textbf{91.3}&\textbf{90.5}&\textbf{90.9}
\\
SENet-18~\cite{hu2019squeeze} & 18 & 11.8M &\textbf{92.9}&91.2&90.3&90.8\\
CBAM-ResNet-18~\cite{woo2018cbam} & 18 & 11.3M & 92.3&90.0&89.8 &89.8 %
\\

FPN-ResNet-18~\cite{lin2017feature} & 18 & 13.3M & 91.9&89.4&88.5&89.0\\
\hline
\textbf{APN-CSCA-18 (ours)}&18&13.8M&93.5&91.7&91.2&91.4\\
\textbf{APN-CSCA-$\theta$-18 (ours)}&18&13.8M&93.5&92.2&91.7&91.9\\
\textbf{APN-CSCA-$\theta^{+}$-18 (ours)}&18&13.8M&\textbf{\color{red}93.7}&\textbf{\color{red}92.7}&\textbf{\color{red}92.2}&\textbf{\color{red}92.4}\\
\hline
pre-act ResNet-34~\cite{he2016identity} & 34 & 21.3M  &92.6&91.2 & 90.1 & 90.6\\
SENet-34~\cite{hu2019squeeze} & 34 & 22.0M &\textbf{93.5}&\textbf{91.6}&\textbf{91.2}&\textbf{91.4}
\\
\hline
\hline
ResNeXt-50~\cite{xie2017aggregated} & 50 & 23.2M & 92.5&89.3&88.5&88.9
\\
RAN-56~\cite{wang2017residual} & 56 & 30.0M & 94.0& \textbf{92.3} & \textbf{91.9} &\textbf{92.1}\\
CBAM-ResNet-50~\cite{woo2018cbam} & 50 & 26.3M & 91.3 & 90.1 &90.0 & 90.1\\
FPN-ResNet-50~\cite{lin2017feature} & 50 & 26.4M & \textbf{94.2}& 92.1&91.2&91.6\\
\hline
\textbf{APN-CSCA-50 (ours)} & 50 & 26.9M & \textbf{\color{red}94.9}&\textbf{\color{red}93.1}&\textbf{\color{red}93.3}&\textbf{\color{red}93.2}
\\
\bottomrule
\end{tabular}
\label{tab_other_models}
\end{table}
Currently, there are lots of excellent deep neural network methods proposed. In order to illustrate the superiority of our method, some experiments are conducted to make comparison between these general methods and our proposed methods, where our methods take pre-act ResNet~\cite{he2016identity} as the backbone network. Furthermore, we also compare our methods with FPN that uses the recently proposed attention models.
\\
\textbf{Comparison with recently proposed methods.}  The experimental results are shown in Table  \ref{tab_other_models}, where all methods are grouped. The optimal results in each group are bold, and the best records are highlighted in red among all methods with similar parameters.
It can be found that the proposed methods obtain the best results. For example, compared with the best representative method SENet-18, APN-CSCA-$\theta^{\text{+}}$-18 obtains the improvement of the performance by 0.8\% of top-1 accuracy, although it has more 2.0M parameters. In order to investigate whether the improvements of the performance just results from the larger parameters, we compare SENet-34 with our APN-CSCA-18. It can be seen that our model with 13.8M parameters still outperforms SENet-34 with 22.0M parameters. For larger models, APN-CSCA-50 outperforms RAN-56 by 0.9\% of top-1, 0.8\% of precision, 1.4\% of recall and 0.9\% of F1 but with the fewer parameters than RAN-56.

More importantly, our methods obtain the better results with the fewer parameters than the larger models such as VGG and WRN. For example, although our methods only achieve improvements by 0.8\% and 0.7\% of top-1 accuracy compared with SENet-18 and FPN-56 respectively, our methods, taking pre-act ResNet as the backbone, consistently outperform all compared methods for all metrics. It is much expected that we can obtain the further improvements if we replace our backbone with the better networks.
\begin{table}[t]
\centering
\tiny
\caption{Comparison of our methods with FPN with recently proposed attention models on CNH-98. Our methods are shown in \textbf{bold}.}
\begin{tabular}{l c c c c c c c}
\toprule
\multirow{2}{*}{\textbf{Model (\# Depth)}} &\multirow{2}{*}{\textbf{Params}} & \textbf{FLOPs} & \textbf{top-1}&\multirow{2}{*}{\textbf{P}}&\multirow{2}{*}{\textbf{R}}&\multirow{2}{*}{\textbf{F1}}\\
~&~&($\times 10^9$)&(\%)&~&~&~&~\\
\hline
FPN + SE~\cite{hu2019squeeze} (18) & 13.4M&4.34 & 92.5 & 90.4 &90.4 &90.3 \\
FPN + CBAM~\cite{woo2018cbam} (18) & 13.4M&4.34 & 92.1 & 89.1&89.5&89.3\\
FPN + SSCA~\cite{Li2018Harmonious} (18) & 13.7M&4.88 & \textbf{93.1} & \textbf{90.9} & \textbf{90.7} & \textbf{90.8}\\
\hline
\textbf{APN-CSCA (18)} & 13.8M&4.88 & 93.5 &91.7&91.2&91.4\\
\textbf{APN-CSCA-$\theta$ (18)}& 13.8M&4.90 & 93.5 & 92.2 & 91.7 & 91.9\\
\textbf{APN-CSCA-$\theta^+$ (18)}& 13.8M&4.90 & \textbf{\color{red}{93.7}} & \textbf{\color{red}92.7} & \textbf{\color{red}92.2} &\textbf{\color{red}92.4}\\
\midrule
\midrule
FPN + SE~\cite{hu2019squeeze} (34) & 23.5M&6.19 & \textbf{93.3} & \textbf{91.2} & \textbf{91.4} & \textbf{91.1} \\
FPN + CBAM~\cite{woo2018cbam} (34) & 23.5M&6.19 & 92.5&91.0&90.3&90.6
\\
FPN + SSCA~\cite{Li2018Harmonious} (34) & 23.9M &6.73& 93.2 & 90.9 & \textbf{91.4} & 90.9\\
\hline
\textbf{APN-CSCA (34)} & 23.9M&6.73 & 93.8 & 92.2 & 91.6 & 91.9\\
\textbf{APN-CSCA-$\theta$ (34)} & 23.9M&6.75 & 93.9 & 92.7& 92.3 &92.5\\
\textbf{APN-CSCA-$\theta^+$ (34)} & 23.9M&6.75 & \textbf{\color{red}{94.0}} & \textbf{\color{red}92.9} &\textbf{\color{red}92.7}& \textbf{\color{red}92.8}\\
\bottomrule
\end{tabular}
\label{tab:fpn+attention}
\end{table}
Note that the effectiveness of directly using FPN to model is not obvious, while our pyramid network with attention module can achieve the apparent improvement. These show that with the help of the proposed attention model, the pyramid network can nicely fuse features with different scales to  solve the multi-granularity recognition task. Because FPN has not the attention mechanism, it cannot adjust adaptively the feature information based on the different granularity so that it hinders the pyramid structure from handling the multi-granularity recognition task.\\
\textbf{Comparision with recently proposed attention models.} In order to validate the superiority of our attention module in the pyramid network, we compare our model APN with FPN with the powerful attention module SE~\cite{hu2019squeeze} (FPN+SE), CBAM~\cite{woo2018cbam} (FPN+CBAM) and SSCA~\cite{Li2018Harmonious} (FPN+SSCA). The experimental results are presented in Table  \ref{tab:fpn+attention}. It can be observed that our methods outperform FPN with any attention model. Although FPN use SSCA as the attention model to obtain the better performance than it uses the other attention models, our method with 18 layers outperforms it (FPN+SSCA) with the same layers by 0.6\% of top-1 accuracy, 1.8\% of precision, 1.5\% of recall, and 1.6\% of F1. For the larger models with 34 layers, our models still have the obvious superiority.
\subsection{Multi-granularity Evaluation of Proposed Method}
\begin{table*}
\center
\scriptsize
\caption{\textbf{The accuracy (\%)} of each model are evaluated by fine-grained metric on each species for 98 classes and coarse-grained metric (\textit{c-Acc}) on the whole datasets for 8 species, where \textit{Total} refers to the accuracy (\%) for 98 classes on the whole dataset. Our methods are shown in \textbf{bold}.}
\begin{tabular}{l c c c c c c c c c c}
\toprule
\multirow{2}{*}{\textbf{Model}} & \multirow{2}{*}{\textbf{Total}} & \textbf{Fruits} & \textbf{Rhi} & \multirow{2}{*}{\textbf{Flowers}} & \multirow{2}{*}{\textbf{Bark}} & \textbf{Thallo} & \textbf{Whole} & \multirow{2}{*}{\textbf{Leaves}} & \multirow{2}{*}{\textbf{Resin}} & \multirow{2}{*}{\textbf{c-Acc(\%)}}\\
~&~&\textbf{\&Seeds}&\textbf{-zome}&~&~&\textbf{-phyte}&\textbf{Herbs}\\
\midrule
ResNet-18~\cite{he2016identity} & 91.7 & 91.4 & 89.6 & 95.5 & 95.3 & 92.3 & 89.6 & \textbf{96.7} & \textbf{100.0} & 87.8\\
FPN-Res18~\cite{lin2017feature} & 91.9 &90.7 & 90.6 & 96.7 & \textbf{96.5} & \textbf{95.6} & 88.1 & \textbf{96.7} & 90.9 & 94.4\\
\midrule
\textbf{APN-CA-18} & 92.9 & 91.9 & \textbf{92.5} & \textbf{98.0} & \textbf{96.5} & 92.3 & \textbf{91.0} & 93.3 & 90.9 & 95.3\\
\textbf{APN-CSCA-18} & \textbf{93.5} & \textbf{93.5} & 90.6 & \textbf{98.0} & \textbf{96.5} & 90.1 & \textbf{91.0} & \textbf{96.7} & 90.9 & \textbf{95.7}\\
\bottomrule
\end{tabular}
\label{tab_cate}
\end{table*}
In order to validate that our models have the ability to solve the problem of inconsistent recognition granularity, a series of experiments are conducted and evaluated with multi-granularity metrics. The experimental results are shown in Table.\ref{tab_cate}, where the column \textit{Total} indicates the best results obtained in experiments for 98 classes on CNH-98. For the more fine-grained evaluation, we evaluated our models and baseline~\cite{he2016identity} on each species for 98 classes. As shown in Table \ref{tab_cate}, our methods obtain the best records in the 6 of 8 species on CNH-98. More importantly, performances of our approaches are fairly more stable and robust for various species, such as all accuracy of our every method on each species are over 90\% for CNH-98. As for the worse results like Resin, we argue that there is actually not a far distance of performances between baseline and our models, because the number of samples in these species is extremely small as shown in Figure  \ref{fig_pie_bar}. For the more coarse-grained evaluation, we also measured our approaches and baseline~\cite{he2016identity} on the whole CNH-98 for eight species, whose results are denoted by \textit{c-Acc} in Table. \ref{tab_cate}. The difference between \textit{Total} and \textit{c-Acc} is that \textit{Total} means using 98 class as labels for CNH-98, while \textit{c-Acc} uses 8 species as labels such as \textit{Bark} in Table. \ref{tab_herbs_details}. It can be found that our model APN-CA still maintains an obvious advantage and APN-CSCA obtains the further improvements. These results indicate that our methods can nicely accomplish multi-granularity recognition tasks such as Chinese herbal recognition.
\subsection{Visualization Analysis}
\begin{figure*}
    \centering
    \includegraphics[scale=0.7]{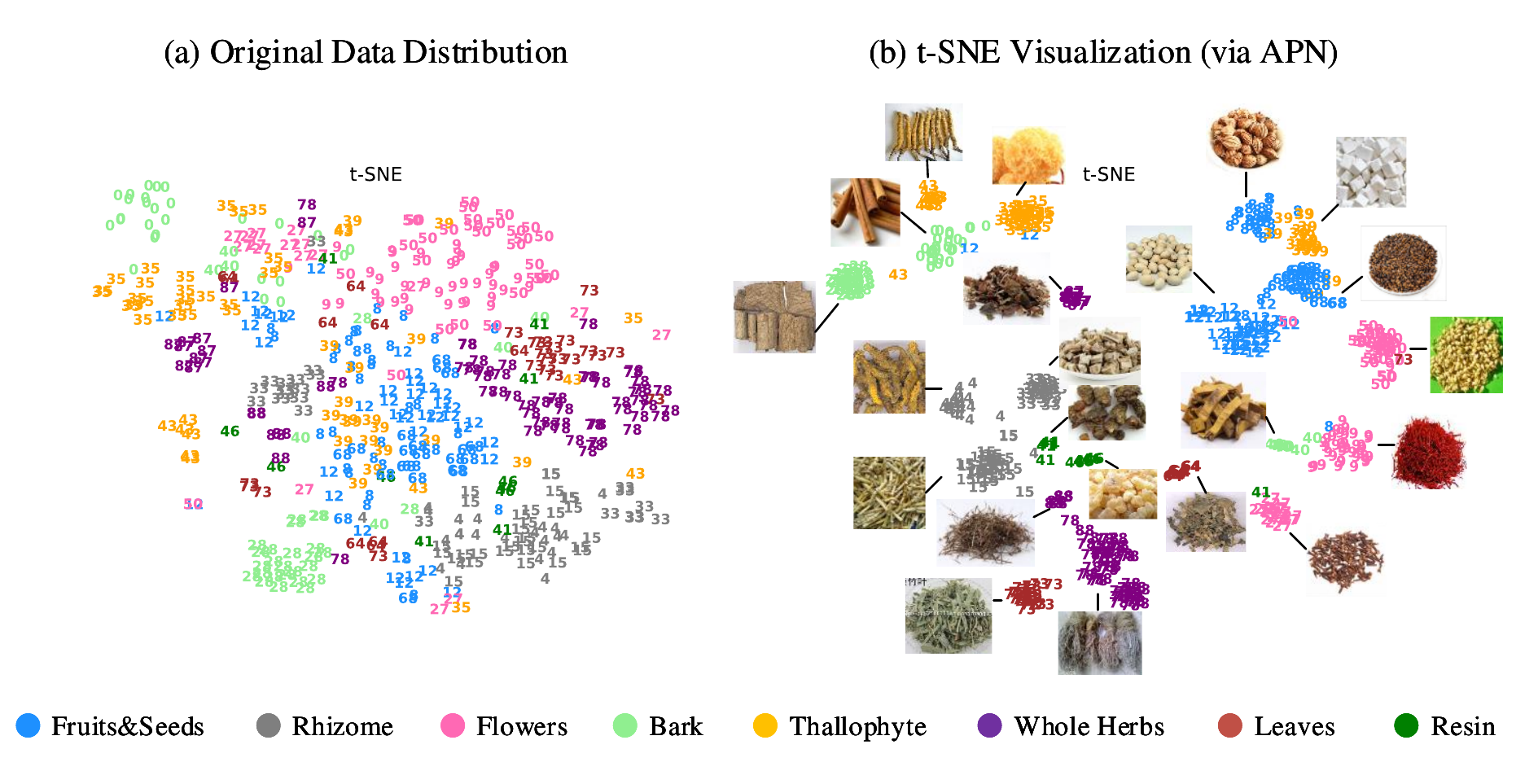}
    \caption{t-SNE visualization of features from randomly selected classes from each species. The color represents different species labels and each number means different class labels of each species. Our proposed method provides the better inter-species discrimination and intra-species discrimination.}
    \label{fig:tsne}
\end{figure*}
In order to confirm our motivation and clearly explain the effectiveness of our proposed methods, the features distribution and the intermediate activation of attention modules in our method are visually analyzed. As shown in Figure  \ref{fig:tsne}, we adopt t-SNE~\cite{dermaaten2008visualizing} to visualize the original data distribution before training and distribution of features extracted by the proposed method (APN) trained on CNH-98. Moreover, we statistics activation values of CA from two pathway in the process of fusing features for four levels and their SCA masks. For a fair comparison, we have normalized the activation values of CA modules into [0, 1].\\
\textbf{(1) t-SNE Visualization.} Because herbs belonging to different species are easier to be recognized while those belonging to the same species are so similar that they should be classified with the more fine-grained cues, herbs recognition can be regarded as a multi-granularity task. In order to validate this, the data distribution of CNH-98 via t-SNE are visualized in Figure \ref{fig:tsne}. Considering too many classes, three classes are randomly selected from each species to be visualized following the practice in ~\cite{zhu2018generative}. In Figure \ref{fig:tsne} (a), features extracted by ResNet~\cite{he2016identity} pretrained on ImageNet~\cite{deng2009imagenet} but without trained on CNH-98 are visualized to show the original data distribution. It can be observed that there are clear boundaries between samples of different species (different color), while samples of different classes in intra-species are mixed together. This indicates that inter-species samples are distinguishable and the intra-species are confused. Furthermore, we also adopt t-SNE to visualize features modeled by APN-CSCA-$\theta^+$ trained on CNH-98 and results are shown in Figure \ref{fig:tsne} (b). Obviously, all classes are much separated. More interestingly, it can be observed that samples of intra-species classes are grouped much closer and those from different species are more separated. These observation indicate our methods not only could separate those from different species but also make intra-species classes more distinguished, validating the proposed methods.
\\
\begin{figure*}
\centering
\includegraphics[scale=0.42]{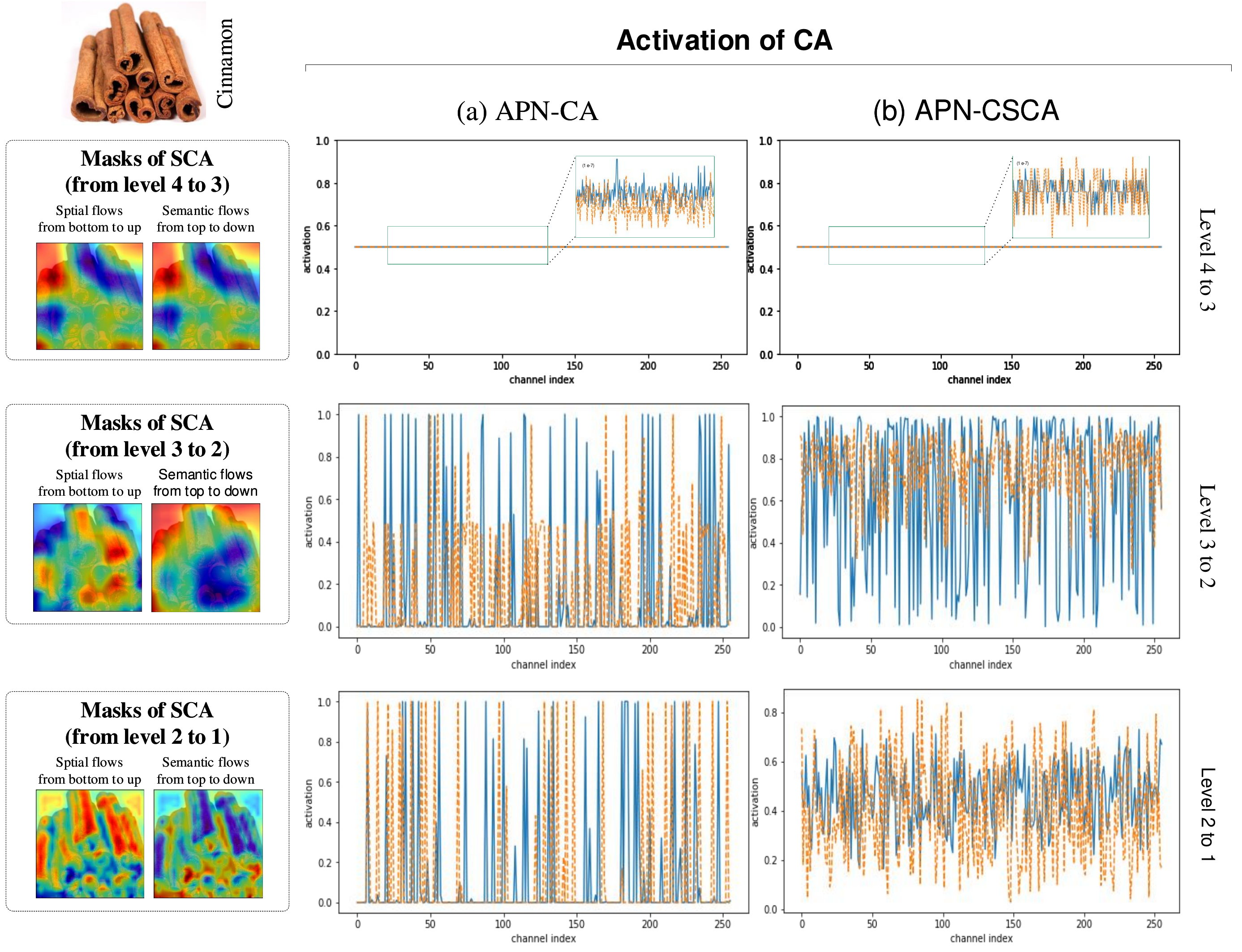}
\caption{Visualization of attention activation for example. In \textit{Masks of SCA (left)}, the color of regions tends to be {\color{red}red}, indicating the higher activation values, and vice versa, to be {\color{blue}blue}. In \textit{Activation of CA (right)}, the {\color[RGB]{30,144,255}blue} lines refer to the activation of spatial flow via lateral connection $\mathbf{s}_l^{spa}$, while the {\color[RGB]{255,127,36}orange} lines mean the other ones $\mathbf{s}_l^{sem}$.}
\label{fig_features1}
\end{figure*}
\textbf{(2) Visualization of Attention Activation.}
As shown in masks of SCA in Figure  \ref{fig_features1} (left), SCA modules of deeper levels tend to focus on the global information. Thus they always activate continuous regions covering the most of target. On the contrary, the shallower ones pay more attention to local details and activate some fragmentary and small regions. These phenomenons validate our motivation for designing SCA to fuse the global and local information. More significantly, the masks of spatial flow and semantic flow show an obvious complementary relationship except for the deepest level. As for the masks of the deepest level, we argue that both of them are extracted by deep layers and have a similar focus. It can also be observed from Figure  \ref{fig_features1} (right) that the activation of CA modules illustrates their dynamical adjustment process. It is obvious that the activation values of level 4 to level 3 are always at about 0.5 and fluctuate slightly (about 1e-7), providing additional evidence for the above argument that deep layers have similar concerns. However, when going shallower, activation of semantics flow shows more densely, and spatial flow shows more sparsely. Furthermore, the CA activation of APN-CSCA is more cautious than that of APN-CA, as features of APN-CSCA on a variety of channels are less suppressed, their activation values are more average, and they fluctuate more stably than that of APN-CA. In such case, SCA modules take the part of the responsibility for modelling, and thus adjustments of CA are less aggressive.\\
\indent
It can be concluded that the attentional activation values of CA and SCA modules are very vigorous and distinguished. They can dynamically and adaptively adjust feature information of different levels and model their complementary relationship for effective features fusion.
\section{Conclusion}
As currently there is no public Chinese herbal image dataset available at present, it is hard to apply machine learning methods to perform Chinese herbal recognition. This paper constructs a new Chinese herbal image dataset for recognition, which can be taken as benchmark datasets to verify any machine learning method. Subsequently, a novel Attentional Pyramid Networks (APN) is proposed that presents a better solution to problems of multi-scale feature fusion. In this new work, Competitive Attention, Spatial Collaborative Attention, and their parametric variants are newly proposed and then applied to model the relationship of features extracted by various layers of APN, so that it can adaptively model each Chinese herbal image with different feature scales. It is validated through experiments that APN can be more efficient to establish a new framework for Chinese herbal recognition. In the future, we will continue to collect more data to support the research of machine learning methods. At the same time, we will take into account Chinese herbal knowledge as the heuristic knowledge to develop new machine learning approaches for Chinese herbal recognition.

\bibliographystyle{cas-model2-names}

\bibliography{references}


\end{document}